\PassOptionsToPackage{dvipsnames}{xcolor}
\RequirePackage{xr-hyper}
\documentclass[numsec,webpdf,modern,medium]{oup-authoring-template}%

\hypersetup{
   colorlinks=true,
  linkcolor={blue},
  filecolor={Maroon},
  citecolor={blue},
  urlcolor={blue},
  pdfcreator={LaTeX via pandoc}}

\usepackage[T1]{fontenc}
\usepackage{graphicx}
\usepackage{amsmath}
\usepackage{amssymb}
\usepackage{adjustbox}
\usepackage{pifont}
\usepackage{makecell}
\newcommand{\cmark}{\ding{51}}
\newcommand{\xmark}{\ding{55}}
\usepackage{booktabs}
\usepackage{longtable,booktabs,array}
\usepackage{calc} 
\usepackage{etoolbox}
\usepackage{lscape}
\usepackage{longtable}
\usepackage{booktabs}
\newcommand{\ms}[2]{%
  \substack{#1\\[0.25ex]\scriptstyle \pm #2}%
}

\newcommand{\bms}[2]{%
  \substack{\mathbf{#1}\\[0.25ex] \scriptstyle \mathbf{\pm #2}}%
}

\newcommand{\ubms}[2]{%
  \substack{\underline{\mathbf{#1}}\\[0.25ex]
            \scriptstyle \underline{\mathbf{\pm #2}}}%
}

\newcommand{\xms}[2]{%
  \ensuremath{\makecell{#1 \\[-0.7ex] \scriptstyle \pm #2}}%
}

\newcommand{\xbms}[2]{%
  \ensuremath{\makecell{\mathbf{#1} \\[-0.7ex] \scriptstyle \mathbf{\pm #2}}}%
}

\newcommand{\xubms}[2]{%
  \ensuremath{\makecell{\underline{\mathbf{#1}} \\[-0.7ex] \scriptstyle \underline{\mathbf{\pm #2}}}}%
}
\usepackage{multirow}
\usepackage{rotating}

\def\bSig\mathbf{\Sigma}

\onecolumn 

\graphicspath{{Fig/}}
\newcommand{\societylogo}{}

\theoremstyle{thmstyleone}%
\newtheorem{theorem}{Theorem}
\newtheorem{proposition}{Proposition}%
\newtheorem{lemma}[theorem]{Lemma}%
\newtheorem{corollary}[theorem]{Corollary}%
\theoremstyle{thmstyletwo}%
\newtheorem{remark}{Remark}%
\theoremstyle{thmstylethree}%

\begin{document}

\journaltitle{Preprint}
\copyrightyear{2026}

\firstpage{1}


\title[EnSurv]{EnSurv: Deep Generative Survival Modeling with Application to Kidney Graft Outcomes}

\author[1,2]{Rajdeep Pathak}
\author[3]{Tamalika Koley}
\author[4]{Shreyashi Basak}
\author[5]{Dipankar Bandyopadhyay}
\author[1, 2, $\ast$]{Tanujit Chakraborty\ORCID{0000-0002-3479-2187}}

\address[1]{\orgdiv{Sorbonne Center for Artificial Intelligence}, \orgname{Sorbonne Universit\'e}, \orgaddress{\state{Paris}, \country{France}}}
\address[2]{\orgdiv{SAFIR}, \orgname{Sorbonne University Abu Dhabi}, \orgaddress{\country{United Arab Emirates}}}
\address[3]{\orgdiv{Decision Sciences Area}, \orgname{Indian Institute of Management Lucknow}, \orgaddress{\state{Uttar Pradesh}, \country{India}}}
\address[4]{\orgdiv{Department of Mathematics}, \orgname{BITS Pilani K K Birla Goa Campus}, \orgaddress{\state{Goa}, \country{India}}}
\address[5]{\orgdiv{Department of Biostatistics}, \orgname{Virginia Commonwealth University}, \orgaddress{\state{Richmond, Virginia}, \postcode{23298-0032} \country{USA}}}

\corresp[$\ast$]{Corresponding author. \href{email:tanujit.chakraborty@sorbonne.ae}{tanujit.chakraborty@sorbonne.ae}}






\abstract{Kidney transplantation is the optimal treatment for life-threatening end-stage renal disease, providing significant survival advantages over dialysis. Given the severe shortage of donor organs, accurately predicting the time to premature graft failure is an essential clinical task for optimizing organ allocation and tailoring individualized immunosuppressive therapies. This study is motivated by a comprehensive United Network of Organ Sharing (UNOS) dataset containing 243,443 kidney transplant records with complex multi-region structures and high right-censoring rates. We propose EnSurv (survival engression), a deep generative survival framework designed to capture complex, non-linear clinical covariate interactions for modeling time to graft failure. Within a pre-additive noise model (pre-ANM) family, EnSurv optimizes an energy score augmented with pooled Kaplan-Meier tail completion to handle right-censoring. We identify the pooled-completion target of the objective, establish consistency for its best approximation within the pre-ANM family, and give conditions and error bounds for its relation to true conditional graft survival. EnSurv also provides global explainability through summaries of fitted feature weights and incorporates domain constraints through established clinical directions using a soft gradient penalty. Extensive simulation experiments were conducted to evaluate the performance of the proposed method in comparison with existing methods. Application to the UNOS cohort demonstrates that EnSurv effectively models individual and region-level effects, consistently outperforming baselines across evaluation metrics encompassing discrimination, calibration, and accuracy. Code is available at \url{https://github.com/PyCoder913/EnSurv} and the \href{https://pypi.org/project/ensurv/}{\texttt{ensurv}} Python package provides an implementation of our approach.}






\maketitle


\section{Introduction}
End-stage renal disease is a life-threatening condition for which kidney transplantation remains the optimal treatment, offering significant survival advantages over dialysis \citep{yoo2016superior}. However, because donor organs are a severely limited resource, optimizing the allocation of available kidneys is a critical public health priority. A primary challenge in this allocation process is that the rate of premature graft loss remains undesirably high \citep{perl2014kidney}. Pre-transplantation, an accurate estimate of a graft's potential longevity is the most relevant information for patients and clinicians when deciding whether to accept a specific kidney offer. In contrary, post-transplantation, reliable survival predictions at an individual level help clinicians tailor patient-specific interventions, such as adjusting the intensity of immunosuppressive therapies to proactively extend graft survival. Understanding the interactions of donor and recipient related clinical factors to predict time to graft failure is therefore an essential clinical task for improving outcomes in both pre- and post-transplant settings.


Predicting graft longevity falls under the framework of survival analysis, which models time-to-event outcomes. A patient may still have a functioning graft when the study ends, or may be lost to follow-up before graft failure occurs. For such a patient, the exact time of graft failure is unknown; the only information available is that the graft remained functional up to their last observed follow-up time. Such an observation is said to be right-censored. 
Historically, classical survival frameworks, such as Cox proportional hazards (CoxPH) \citep{cox1972regression} and accelerated failure time (AFT) \citep{wei1992accelerated} models have been the gold standard for analyzing censored data. Indeed, established clinical scoring systems for predicting post-transplant kidney outcomes, including the estimated post transplant survival score \citep{clayton2014external}, the recipient risk score \citep{baskin2007recipient}, and life years from transplant \citep{wolfe2009predictability} were largely developed using CoxPH formulations. While these classical models are highly interpretable, they solely rely on strict structural assumptions, such as proportional hazards or predefined parametric distributions. These assumptions can be restrictive and often fail to capture the complex, non-linear, and high-dimensional clinical covariate interactions present in modern transplant databases.

Our study is motivated by a kidney transplantation dataset derived from the United Network of Organ Sharing \citep[UNOS;][]{dharnidharka2005bloc}. The dataset considered in this study includes time to graft failure for 243,443 kidney transplantations performed from October 1987 to June 2016, with an event right-censoring rate of approximately $60\%$. Beyond its sheer volume, this data exhibits a complex hierarchical and multi-regional structure, as each transplant is performed at specific centers nested within 11 distinct geographical regions. Numerous prior studies investigating kidney graft failure and simultaneous organ transplants have extensively adjusted for or matched on the clinical covariates, while implicitly 
ignoring region-induced effects \citep{gupta2017long, kamal2020impact}. Given the high-stakes nature of organ allocation or post-transplant treatments, accurately estimating graft survival requires a framework that systematically captures both individual donor-recipient profiles and these region-level effects. Furthermore, an effective clinical tool must go beyond black-box predictions by providing inherent explainability to rank early failure risk factors, while integrating domain knowledge regarding which covariates are clinically established to accelerate or delay graft loss.

Given the vast sample size and the high dimensionality of modern electronic health record datasets, deep learning approaches for survival analysis \citep{katzman2018deepsurv, lee2018deephit} have emerged as powerful tools capable of capturing complex, non-linear clinical relationships. However, while standard neural survival architectures excel at predictive accuracy in high-dimensional spaces, they function largely as black boxes, lacking both the inherent explainability required to rank clinical risk factors and the architectural mechanisms necessary to integrate established domain knowledge. 
To this end, we introduce EnSurv, a deep generative framework for prediction of kidney graft survival outcomes. EnSurv employs a pre-additive noise architecture to generate conditional log-failure times and is trained using a combination of energy score loss used in engression framework \citep{gneiting2007strictly, shen2025engression, pathak2026engression}, and a pooled Kaplan-Meier (KM) tail-completion term to accommodate right-censored observations in the UNOS registry. In contrary to Censored engression \citep{jonkers2026proper}, which scores generated outcomes after a censoring transformation and studies propriety on an identifiable region, EnSurv targets a horizon-restricted working outcome distribution constructed through pooled KM completion. This approach avoids separately estimating covariate-dependent censoring distributions by employing a shared residual-tail estimate, enabling EnSurv to learn patient-specific predictive distributions and generate individualized graft survival curves up to a prespecified horizon. It is generative since for a fixed patient covariate profile, independent noise draws passed through the fitted network yield log-time samples from this learned distribution. Unlike standard deep learning models, EnSurv provides built-in global explainability to identify the primary drivers of early graft loss. It ranks feature importance by jointly tracking the explicit weights of a linear AFT-like component and the weight norms of the non-linear dense layers. Moreover, EnSurv integrates clinical domain knowledge via gradient penalization during backpropagation. This allows for encouraging specified directional associations during training, such as learning that a higher kidney donor risk index (KDRI) or presence of hepatitis C virus (HCV) in the donor correlates with reduced graft survival \citep{slabiak2023association}. 

Our theoretical analysis establishes that EnSurv's objective is strictly proper for the conditional outcome law induced by pooled KM completion. We characterize the conditional distribution induced by pooled completion, and establish uniform convergence of the objective and consistency of approximate minimizers for the best attainable pooled-target risk. We also quantify pooling distortion and state the additional conditions under which the true restricted law can be recovered. 
We use simulations to empirically examine how the KM approximation and empirical risk approach their population counterparts as the sample size grows, and compare EnSurv’s predictive performance with that of existing survival methods across a range of data-generating mechanisms and censoring levels. Finally, application to the UNOS cohort establishes that EnSurv consistently outperforms state-of-the-art survival frameworks, including survival foundation model, in predicting graft failure time across a broad suite of ten evaluation metrics encompassing discrimination, calibration, and accuracy. Hence, EnSurv serves as a comprehensive toolkit for predicting kidney graft survival, functioning as a deep generative architecture that natively integrates established domain knowledge and robust explainability. Our main contributions are four-fold:
\begin{enumerate}
    \item We introduce EnSurv, a deep generative model that samples graft failure times for a given donor–recipient profile and yields individualized survival curves up to a prespecified horizon. 
    \item We combine an energy score with pooled Kaplan–Meier completion of censored records, characterize the resulting conditional distribution, and establish risk consistency for this target. Extensive simulation experiments are conducted to validate the theoretical results as well as evaluate predictive performance against existing methods.
    \item  EnSurv makes its predictions open to clinical interpretation. Signed coefficients describe associations in the linear component, while weights along nonlinear network paths help rank the importance of patient and regional covariates. A gradient penalty allows established clinical knowledge to guide the direction of selected associations during training, thereby encouraging domain constraints.
    \item Across approximately 240,000 UNOS kidney transplants, EnSurv achieved the highest mean concordance among the methods evaluated, both before transplantation (0.609) and after transplantation (0.663). Explainability analysis post transplant revealed that dialysis in the first week after transplant, higher creatinine at discharge, and higher KDRI were among the strongest markers of shorter predicted graft survival.
\end{enumerate}

The remainder of this article is organized as follows. Section~\ref{sec:UNOS} describes the dataset. Section~\ref{sec:methodology} introduces the proposed framework and the model architecture, and Section~\ref{sec:Theoretical} studies theoretical properties. Simulation experiments and real data analyses are presented in Sections \ref{sec:simulations} and \ref{sec:real_data}, respectively. Finally, Section~\ref{sec:conclusion} concludes the work, discussing limitations and potential future directions.

\section{Motivating Example: UNOS Dataset}\label{sec:UNOS}
\begin{figure}[!t]
    \centering
    \includegraphics[width=0.7\textwidth]{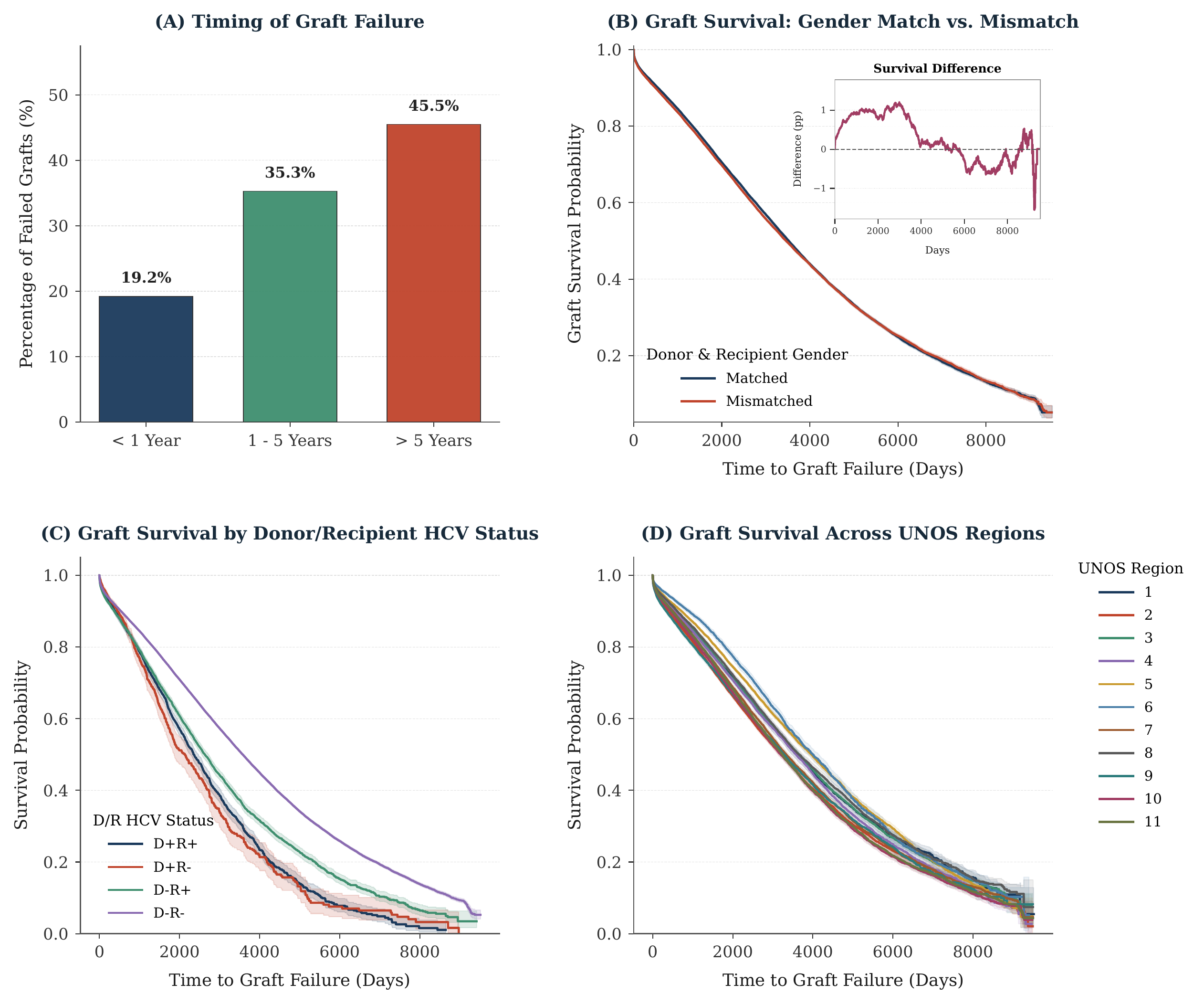}
    \caption{Composite analysis of kidney graft outcomes from the UNOS registry: (A) distribution of failure times; and survival probabilities stratified by (B) donor-recipient gender concordance, (C) HCV serological status, and (D) UNOS geographic region.}
    \label{fig:eda_analysis}
\end{figure}
At the patient level, the available prognostic covariates include demographic attributes (age, gender, and race) alongside baseline clinical variables. Key clinical indicators include hospital length of stay (LOS) post transplant, panel reactive antibodies (PRA), donor and recipient blood types, discharge creatinine levels, KDRI, and immunosuppression strategies. Preliminary analysis shows that approximately $60\%$ of the donors and recipients are male, and about $26\%$ of the patients required dialysis within the first week of transplant. Among patients experiencing graft failure, the timing of failure varied, with 19.2\% failing within the first year (short-term), 35.3\% between 1 and 5 years, and 45.5\% failing after 5 years. Comparing donor-recipient gender matching, 52.9\% of the transplants were gender-matched, while 47.1\% were mismatched (female receiving male kidney or vice versa). The survival trajectories between these groups exhibited a statistically significant difference (log-rank $p < 0.001$), with gender-matched transplants showing a higher initial survival probability. Stratification by HCV serological status of donors and recipients demonstrated a statistically significant impact on overall graft survival ($p <0.001$). The D-R- group (both donor and recipient HCV negative) exhibited the highest survival probability over time, whereas the HCV positive donor (D+) with HCV negative recipient (R-) pairing had the lowest survival. These primary survival analyses for the cohort are summarized in Panels (A), (B), and (C), respectively, in Figure~\ref{fig:eda_analysis}.

The UNOS registry inherently possesses a multi-regional nature, where each patient comes from one of 11 distinct Organ Procurement and Transplantation Network (OPTN) regions. A multivariate log-rank test on the survival curves of patients stratified by region indicated a statistically significant variance in graft survival rates ($p<0.001$; Figure~\ref{fig:eda_analysis}(D)). Specifically, the estimated 10-year restricted mean graft survival was highest in
Region~6 and lowest in Region~2. Hence, ignoring region-specific effects might result in failing to account for unobserved region-level heterogeneity driven by systemic variations in surgical protocols, region volume, and post-operative care. Detailed descriptive statistics and clinical definitions for all covariates are provided in Table~\ref{table:data_explanation} (supplementary Section~\ref{appendix:tables_figures}).

\section{Methodology}\label{sec:methodology}
\subsection{Problem Formulation}
Let $T>0$ be time to graft failure, $C$ the right-censoring time (possibly infinite), and $X\in\mathcal X\subseteq\mathbb R^d$ the covariate vector, with $\mathcal X$ a Borel set. We observe $n$ independent, identically distributed copies $(Y_i,\delta_i,X_i)$ of
\[
Y=\min(T,C),\qquad \delta=\mathbf1\{T\le C\},
\]
where $Y$ denotes the time to graft failure or censoring (whichever occured first) and $\delta$ is the graft failure indicator.
Fix a clinically chosen horizon $\tau$ and a positive lower bound $c_0$ on censoring times, with $0<c_0<\tau<\infty$. Define
\begin{equation*}
\begin{gathered}
Z=\log T,\qquad a=\log\tau,\qquad Z_\tau=\min(Z,a),\\
A=\log\min(Y,\tau),\qquad D=1-(1-\delta)\mathbf1\{Y<\tau\}.
\end{gathered}
\end{equation*}
Thus, $D=1$ means that $Z_\tau$ is observed and equals $A$: either graft failure is observed before $\tau$, or follow-up reaches $\tau$. When $D=0$, follow-up was censored before the horizon, i.e., $Y=C<\tau$ and $T>C$, so only $Z_\tau>\log Y$ is known. Follow-up to $\tau$ is sufficient to observe the restricted outcome and a failure at $\tau$ is not required. 
Throughout this study, we consider the following population-level assumptions:
\begin{equation}\label{eq:cond_1}
C\ge c_0\ \text{almost surely},\qquad h_0:=\mathbb P(Y\ge\tau)>0,
\qquad \mathbb E|Z_\tau|<\infty,
\end{equation}
where $h_0$ denotes the probability that an individual's observed follow-up time reaches or exceeds \(\tau\). The positive lower bound on censoring keeps completed log-time tails in the bounded interval $[\log c_0,a]$, and the follow-up condition keeps the pooled risk set positive up to $\tau$. 

Let $P_X$ denote the marginal law of the covariates and $P_x$ be the covariate-conditional law of $Z_\tau$; we write
\[
P_x=\mathcal L(Z_\tau\mid X=x),\qquad F_x(z)=P_x(( -\infty,z]),
\qquad S_T(t\mid x)=\mathbb P(T>t\mid X=x),
\]
where $F_x$ is the conditional cumulative distribution function (CDF) of the horizon-restricted log-failure time and $S_T(t\mid x)$ denotes the conditional survival function representing the probability that an individual with covariates \(x\) experiences no graft failure before or at time \(t\). Throughout, a minus sign after a time argument denotes a left-hand limit, so $f(t-)=\lim_{s\uparrow t}f(s)$, with any other arguments held fixed. All the conditional statements hold for $P_X$-almost every $x$, given that regular conditional distributions exist on the underlying Borel spaces. We additionally consider
\begin{equation}\label{eq:positivity}
T\perp C\mid X,\qquad G(t\mid x):=\mathbb P(C\ge t\mid X=x)>0
\quad(0\le t\le\tau),
\end{equation}
where $G(t\mid x)$ is the conditional probability that an individual with covariates \(x\) remains uncensored through time \(t\). The former condition is conditional independence, which says that given the measured covariates, censoring supplies no further information about the failure time. Positivity in the latter condition permits follow-up throughout the prediction window for almost every profile. Together, they identify the conditional event hazard wherever the conditional risk set is positive:
\[
\frac{\mathbb P(Y\in dt,\delta=1\mid X=x)}{\mathbb P(Y\ge t\mid X=x)}
=\frac{\mathbb P(T\in dt\mid X=x)}{\mathbb P(T\ge t\mid X=x)} = \frac{F_T(dt\mid x)}{S_T(t-\mid x)}.
\]
The conditional hazard that we can estimate from
the censored data is therefore the hazard of the true event and it determines the conditional survival function of $T$ on $[0,\tau]$.

\subsection{Model Formulation}
Let $\nu=(\theta,\phi,\beta)\in\Theta\subseteq\mathbb R^p$ parameterize a predictive family for the horizon-restricted log-time. The generated outcome is defined as
\begin{equation}\label{eq:ensurv_model}
U_\nu(x,\epsilon)=g_\theta\left(x+h_\phi(\epsilon)\right)+\beta^\top x,
\qquad \epsilon\sim P_\epsilon,
\end{equation}
where $h_\phi:\mathbb R^q\to\mathbb R^d$ transforms independent simulation noise $\epsilon$ sampled from $P_\epsilon$ (e.g., Gaussian or uniform), $g_\theta:\mathbb R^d\to\mathbb R$ is a neural network (multilayer perceptron), and $\beta\in\mathbb R^d$ provides a linear pathway.

Let $Q_{\nu,x}=\mathcal L_\epsilon\{U_\nu(x,\epsilon)\}$ denote the predictive law induced by the generated outcome, with CDF $F_{\nu,x}(z)=\mathbb P_\epsilon\{U_\nu(x,\epsilon)\le z\}$. On the original time scale, the model survival function on $(0,\tau)$ is $1-F_{\nu,x}(\log t)$. 
To accommodate right-censored observations ($D=0$), EnSurv imputes unobserved log-times using a pooled KM tail. Let $N(t)=\mathbb P(Y\le t,\delta=1)$ denote the observed graft failure subdistribution and $H(t)=\mathbb P(Y\ge t)$ the at-risk probability. Their pooled hazard and survival functions, respectively, on $[0,\tau)$ are defined as$$\Lambda^\#(t)=\int_{(0,t]}\frac{N(ds)}{H(s)},\qquad S^\#(t)=\prod_{(0,t]}\{1-d\Lambda^\#(s)\}.$$
The superscript $\#$ denotes a working curve constructed from failures and risk sets pooled across the cohort. We consider pooling because a local conditional tail can be unstable late in follow-up, when few patients with similar covariates remain at risk. For a patient censored at $c<\tau$, the pooled curve is used to allocate probability over $(c,\tau]$, so patients censored at the same time share a completion distribution.
Because observed failures represent only a subset of all risk-set removals, $S^\#(t)\ge\mathbb P(Y>t)\ge h_0$ for $t<\tau$. Placing the residual mass $S^\#(\tau-)$ at $\tau$ gives a probability distribution on $(0,\tau]$ with CDF $F^\#$.
For a given censoring time $c\in[c_0,\tau)$, the conditional log-tail law $K_c$ and its expected absolute deviation $\Psi(c,u)$ at $u\in\mathbb R$ are
$$K_c(E)=\frac{\int_{(c,\tau]}\mathbf1\{\log t\in E\}\,dF^\#(t)}{S^\#(c)},\qquad \Psi(c,u)=\int\vert{}z-u\vert{}\,K_c(dz),$$
where $E\subseteq\mathbb R$ is a Borel set. Both $K_c$ and $\Psi$ are constructed from the pooled distribution $F^\#$ and do not depend on the patient covariates. By convention, $\Psi(c,u)=0$ outside $[c_0,\tau)$ to ensure well-defined products when $D=1$.

For any predictive law $Q$ with a finite first moment, model performance is evaluated via the pooled-completion energy score loss:
\begin{equation}\label{eq:pooled_energy_score}
\ell(Q;Y,\delta)=\mathbb E_{U\sim Q}\{D\vert{}A-U\vert{}+(1-D)\Psi(Y,U)\}-\frac12\mathbb E\vert{}U-U'\vert{},
\end{equation}
where $U,U'$ are independent draws from $Q$. The first term measures the accuracy of the generated log-times. When $D=1$, it penalizes the absolute deviation of $U$ from the observed horizon-restricted log-time $A$. When $D=0$, the term $\Psi(Y,U)$ averages this absolute deviation over possible log-failure times drawn from the pooled completion distribution $K_Y$. The second term measures dispersion through the expected absolute distance between two independent draws from $Q$. Subtracting this term rewards separation between generated log-times, counterbalancing the tendency of the accuracy term alone to concentrate predictions near a median. The corresponding population risk is
\begin{equation*}
R(\nu)=\mathbb E\{\ell(Q_{\nu,X};Y,\delta)\}.
\end{equation*}
To characterize the minimizer of this risk, we define an auxiliary pseudo-complete outcome $Z^*$. This outcome equals the observed restricted log-time $A$ when $D=1$, and is drawn conditionally from the tail law $K_Y$ when $D=0$:$$Z^*=A\quad\text{if }D=1,\qquad Z^*\mid(Y,\delta,X)\sim K_Y\quad\text{if }D=0.$$The associated conditional law and CDF are denoted by $P_x^*=\mathcal L(Z^*\mid X=x)$ and $F_x^*(z)=P_x^*((-\infty,z])$, respectively. Because $Z^*=A=Z_\tau$ whenever $D=1$, including observations followed to the horizon without an observed event, and $Z^*\in(\log c_0,a]$ when $D=0$, it satisfies the moment bound
$$\mathbb E(\vert{}Z^*\vert{}\mid X=x)\le\mathbb E(\vert{}Z_\tau\vert{}\mid X=x)+\max\{\vert{}\log c_0\vert{},\vert{}a\vert{}\}<\infty,$$
for almost every $x$. For graft prognosis below $\tau$, $P_x^*$ preserves known outcomes and respects censoring bounds, yielding a covariate-conditional working survival law. The resulting population objective is strictly proper for $P_x^*$, with uniformly consistent tail-loss estimation under the stated assumptions. Section~\ref{sec:Theoretical} bounds the discrepancy between the
pooled-completion law $P_x^*$ and the true conditional law of the
restricted log-failure time, and gives sufficient conditions for the
two laws to agree. Supplementary Section~\ref{appendix:theory} gives an exact
expression for the difference in survival probabilities and bounds
the corresponding difference in restricted mean graft survival.

\subsection{Estimation}\label{sec:estimation}
To estimate the model parameters $\nu$, we draw independent noises $\epsilon_{ij}\sim P_\epsilon$ for each observation $i=1,\ldots,n$ across $m\ge2$ replicates, yielding the generated outputs $u_{ij}(\nu)=U_\nu(X_i,\epsilon_{ij})$. The empirical risk is defined as the Monte Carlo approximation of the energy score:
\begin{equation}\label{eq:emp-loss}
    \begin{split} \widehat R_n(\nu)=\frac1n\sum_{i=1}^n\Bigg[&\frac1m\sum_{j=1}^m \left\{D_i\vert{}A_i-u_{ij}(\nu)\vert{}+(1-D_i)\widehat\Psi_n(Y_i,u_{ij}(\nu))\right\}-\frac1{2m(m-1)}\sum_{j\ne k}\left|u_{ij}(\nu)-u_{ik}(\nu)\right|\Bigg]. \end{split}
\end{equation}

The expected tail deviation $\Psi$ is estimated using an average of normalized bootstrap KM tails. Bootstrapping reduces the sensitivity of the completion loss to sample-specific risk-set fluctuations while preserving the same asymptotic pooled target. We draw $B$ samples of size $n$ with replacement from the training observations. In the $b$-th sample, let $v_{b1}<\cdots<v_{bK_b}$ denote the distinct failure times below the horizon $\tau$, with $d_{b\ell}$ failures and $r_{b\ell}$ subjects at risk at $v_{b\ell}$. The sample KM survival and corresponding probability masses are:

\begin{equation*}
\widehat S_n^{(b)}(t)=\prod_{v_{b\ell}\le t}\left(1-\frac{d_{b\ell}}{r_{b\ell}}\right),\qquad
w_{b\ell}=\widehat S_n^{(b)}(v_{b\ell}-)\frac{d_{b\ell}}{r_{b\ell}}
\quad(t<\tau).
\end{equation*}
Assigning all residual mass $\widehat S_n^{(b)}(\tau-)$ to $\tau$, the estimated conditional deviation for $c_0\le c<\tau$ is defined as
$$\widehat\Psi_n^{(b)}(c,u)= \begin{cases} \frac{1}{\widehat S_n^{(b)}(c)} \left( \sum_{c<v_{b\ell}<\tau}\vert{}\log v_{b\ell}-u\vert{}w_{b\ell} +\vert{}a-u\vert{}\widehat S_n^{(b)}(\tau-) \right), &\widehat S_n^{(b)}(c)>0,\\ \vert{}a-u\vert{},&\widehat S_n^{(b)}(c)=0. \end{cases}$$
The zero-survival fallback captures the point-mass completion at $a=\log\tau$. The final estimator averages these normalized tails:
\begin{equation}\label{eq:emp-psi}
\widehat\Psi_n(c,u)=\frac1B\sum_{b=1}^B\widehat\Psi_n^{(b)}(c,u),
\end{equation}
adopting the same convention as $\Psi$ where $\widehat\Psi_n(c,u)=0$ outside $[c_0,\tau)$.

Given a measurable approximate minimizer $\widehat\nu_n$ of the empirical risk, EnSurv predicts survival for a new covariate profile $x$ using $M$ independent prediction draws $\epsilon_j^{\mathrm{pred}}$. The estimated predictive CDF and corresponding survival function for $0<t<\tau$ are
\begin{equation}\label{eq:prediction-mc}
\begin{split}
\widehat F_{n,M}(z\mid x)&=\frac1M\sum_{j=1}^M
\mathbf1\left\{U_{\widehat\nu_n}\left(x,\epsilon_j^{\mathrm{pred}}\right)\le z\right\},\\
\widehat S_{n,M}(t\mid x)&=1-\widehat F_{n,M}(\log t\mid x),\qquad 0<t<\tau.
\end{split}
\end{equation}

\subsection{Model Architecture}\label{sec:architecture}
The formulation of the EnSurv model is realized through a neural network architecture. An exogenous noise tensor $\epsilon$, sampled from a base distribution such as a Uniform or Gaussian distribution, is injected into the covariates $X\in \mathbb{R}^d$. The noise-augmented representations are subsequently processed by a multi-layer perceptron (MLP), comprising stacked fully connected layers. The layers can be stabilized by one-dimensional batch normalization and regularized via dropout \citep{srivastava2014dropout}. The nonlinear pathway concludes with a terminal linear projection to a scalar value, encapsulating the complex generative function $g_\theta$. Operating in parallel to this generative MLP, a direct linear projection of the covariates $X$ is computed using a zero-initialized, learnable coefficient vector $\beta \in \mathbb{R}^d$. This skip-connection preserves baseline linear effects and provides global explainability by weighing features directly. The final predicted log-event time $\widehat{Z}$ is computed as the superposition of the nonlinear and linear pathways.

The theoretical analysis below concerns the base EnSurv model trained with the pooled-completion energy score. The additional domain-constraint penalty during backpropagation used in the UNOS application are described in Section~\ref{sec:real_data}.


\section{Theoretical Properties}\label{sec:Theoretical}

We first identify the conditional distribution targeted by EnSurv and examine how it relates to graft survival. For a fixed horizon $\tau$ satisfying~\eqref{eq:cond_1}, the pooled-completion energy score is strictly proper for $P_x^*$. Under conditional independent censoring, we bound the discrepancy between $P_x^*$ and the true conditional distribution of the restricted log-failure time and give conditions that determine the direction of this discrepancy. We then turn to estimation. We establish an $O(n^{-1/2})$ bound on the expected uniform error of the estimated completion loss and, under conditions (A1)--(A4), prove uniform convergence of the empirical risk and consistency of approximate minimizers with vanishing optimization error. Supplementary Section~\ref{appendix:theory} develops the supporting results for KM estimation and examines the implications for patient-specific survival probabilities, prediction error, and restricted mean graft survival. All proofs are given in supplementary Section~\ref{appendix:proofs}.


\begin{proposition}\label{prop:target}
Let $\ell$ denote the pooled-completion energy score in \eqref{eq:pooled_energy_score}. For almost every $x$ and any probability law $Q$ with finite first moment, let $r_x(Q)=\mathbb E\{\ell(Q;Y,\delta)\mid X=x\}$ and let $F_Q$ be its CDF. For independent $Z^*\sim P_x^*$ and $U,U'\sim Q$,
\begin{equation}\label{eq:target-risk}
r_x(Q)=\mathbb E|Z^*-U|-\tfrac12\mathbb E|U-U'|,
\end{equation}
and
\begin{equation}\label{eq:energy-identity}
r_x(Q)-r_x(P_x^*)=\int_{\mathbb R}\{F_Q(z)-F_x^*(z)\}^2\,dz\ge0.
\end{equation}
Equality holds if and only if $Q=P_x^*$. If all model risks are finite and the model realizes $P_x^*$ almost everywhere, then every population risk minimizer realizes $P_x^*$ almost everywhere.
\end{proposition} 

Next, we bound the discrepancy between $P_x^*$, the pseudo-complete distribution we are fitting against, and the true unobserved distribution of the restricted
log-failure time $P_x$. For laws $P,Q$ with finite first moments, write
$W_1(P,Q)=\inf\mathbb E|V-U|$ for the 1-Wasserstein distance, where the infimum is over couplings with marginals $P,Q$. On the real line, it equals the integral of the absolute difference between their CDFs. Let $L=\log(\tau/c_0)$ be the largest residual log-time interval length.

\begin{proposition}[Pooling discrepancy and its direction]
\label{prop:pooling}
In addition to \eqref{eq:cond_1}, assume
$T\perp C\mid X$. The following statements hold for $P_X$-almost every
profile $x$ with
$\mathbb E(|Z_\tau|\mid X=x)<\infty$. For $c_0\le c<\tau$ with $S_T(c\mid x)>0$, define the true conditional residual law
\[
R_x(c)=\mathcal L(Z_\tau\mid T>c,X=x)
\]
and
\[
b(x)=
\int_{[c_0,\tau)}
S_T(c\mid x)\,
W_1\{R_x(c),K_c\}\,
P_{C\mid x}(dc).
\]
In case of $S_T(c\mid x)=0$, we define $R_x(c)$ to be the point mass at $a=\log\tau$; its value receives zero weight in the integral.

\begin{enumerate}
\item[(i)] (Pooling discrepancy bound)
The discrepancy satisfies
\begin{equation}\label{eq:pooling-bound}
W_1(P_x^*,P_x)
\le b(x)
\le
L\,\mathbb P(T>C,\ C<\tau\mid X=x).
\end{equation}
Thus, the bound depends on censoring before the horizon, rather than
on the overall censoring probability.

\item[(ii)] (Agreement with the true restricted distribution)
If $R_x(c)=K_c$ for $P_{C\mid x}$-almost every $c\in[c_0,\tau)$ with
$S_T(c\mid x)>0$, then $P_x^*=P_x$.
This is a sufficient condition, and it requires the pooled completion law to agree with the patient's true residual law at the censoring times that receive positive weight.

\item[(iii)] (Direction of pooling discrepancy)
If $K_c$ stochastically dominates $R_x(c)$ at these censoring times,
then $P_x^*$ stochastically dominates $P_x$. Equivalently, $F_x^*(z)\le F_x(z)$ for every $z\in\mathbb R$. Reversing the residual stochastic order reverses the marginal order. A sufficient condition for this dominance is $S_T(c\mid x)>0$ for every $c\in[c_0,\tau)$ and domination of hazard measures:
\[
d\Lambda_x(t)
:=
\frac{\mathbb P(T\in dt\mid X=x)}
     {\mathbb P(T\ge t\mid X=x)}
\ \ge\ d\Lambda^\#(t)
\qquad\text{on }(c_0,\tau).
\]
Here, the inequality is an inequality of measures, allowing both
continuous hazard components and hazard jumps.
Equality of these hazard measures implies the residual agreement
in part~(ii). Reversing their inequality reverses the stochastic order.
\end{enumerate}
\end{proposition}

Hence, pooling error is concentrated among patients censored before the horizon and depends on how their residual prognosis differs from the pooled tail. For a profile with larger hazard measure, pooling can overstate survival; for a lower-risk profile it can understate survival. Utilizing a conditional tail completion instead of the pooled completion law could remove this distortion when consistently estimated at the additional expense of computational overhead, which we posit as a promising future direction.

For fitting the model, impose the following conditions in addition to~\eqref{eq:cond_1}:
\begin{enumerate}
\item[(A1)]\label{assmp:1} $\Theta\subseteq\mathbb R^p$ is nonempty and compact. 
\item[(A2)]\label{assmp:2} $(\nu,x,e)\mapsto U_\nu(x,e)$ is jointly measurable, and $\nu\mapsto U_\nu(X,\epsilon)$ is continuous on $\Theta$ for almost every $(X,\epsilon)$.
\item[(A3)]\label{assmp:3} $\mathbb E[\sup_{\nu\in\Theta}|U_\nu(X,\epsilon)|]<\infty$.
\item[(A4)]\label{assmp:4} The noises in~\eqref{eq:emp-loss} are i.i.d. across $i,j$, independent of the observations and bootstrap draws; $m\ge2$ is fixed. The number of bootstrap replicates is fixed or a deterministic finite integer $B=B_n\ge1$.
\end{enumerate}
Compactness in (A1) ensures population minimizers exist and allows finite-cover arguments. Continuity and measurability in (A2) make the risks well-defined and stable under parameter limits. The uniform moment condition in (A3) supplies an integrable loss envelope; for a fixed ReLU network with bounded parameters it follows, for example, from finite first moments of $\|X\|$ and $\|\epsilon\|$. Finally, independence across observations permits averaging simulation error even with fixed $m$.

\begin{theorem}[Uniform tail-completion consistency]\label{thm:psi}
For the estimator~\eqref{eq:emp-psi} with deterministic finite $B=B_n\ge1$, define
\[
\eta_n=\sup_{c_0\le c<\tau,\,u\in\mathbb R}\left|\widehat\Psi_n(c,u)-\Psi(c,u)\right|.
\]
Then $\mathbb E\eta_n=O\left(n^{-1/2}\right)$ and $\eta_n=O_P\left(n^{-1/2}\right)$. 
\end{theorem}

The above theorem states that replacing the unobservable population residual loss with our finite-sample, bootstrap-aggregated KM approximation introduces an estimation error that uniformly vanishes at $n^{-1/2}$ order, ensuring the asymptotic reliability of the data-driven objective function.

\begin{theorem}[Uniform convergence of the empirical risk]\label{thm:ulln}
Let \(\Delta_n\) denote the uniform deviation of the empirical risk $\widehat R_n$ from the population risk $R$. Under (A1)--(A4), $R(\cdot)$ is finite and continuous on $\Theta$, and
\begin{equation}\label{eq:uniform-loss}
\Delta_n:=\sup_{\nu\in\Theta}\left|\widehat R_n(\nu)-R(\nu)\right|\xrightarrow{P}0.
\end{equation}
\end{theorem}

Hence, as the sample size grows, the empirical risk uniformly converges to the true expected population risk, ensuring that minimizing our empirical objective is asymptotically equivalent to minimizing the population pooled-completion risk over the fixed family. Consequently, we establish in the following theorem that a fit with vanishing empirical optimization error approaches the set of population minimizers and attains, in the limit, the lowest pooled-completion risk available within that family.

\begin{theorem}[Consistency of approximate minimizers]\label{thm:argmin}
Under Theorem~\ref{thm:ulln} and $\widehat{\nu}_n\in \Theta$, suppose the measurable estimate satisfies
\begin{equation}\label{eq:optimization-gap}
\widehat R_n(\widehat\nu_n)\le\inf_{\nu\in\Theta}\widehat R_n(\nu)+r_n,
\qquad r_n\ge0,\quad r_n\xrightarrow{P}0.
\end{equation}
Let $\Theta_0=\operatorname*{arg\,min}_{\nu\in\Theta}R(\nu)$. This set is nonempty and compact, and
\begin{equation}\label{eq:argmin-risk}
0\le R(\widehat\nu_n)-\min_{\nu\in\Theta}R(\nu)\le2\Delta_n+r_n,
\qquad \operatorname{dist}(\widehat\nu_n,\Theta_0)\xrightarrow{P}0,
\end{equation}
where $\operatorname{dist}(\nu,\Theta_0)=\inf_{\mu\in\Theta_0}\|\nu-\mu\|$. Moreover, if $\Theta_0=\{\nu_0\}$ is a singleton, then
$\widehat\nu_n\xrightarrow{P}\nu_0$.
\end{theorem}


\begin{remark}
EnSurv makes use of incomplete UNOS records by retaining each recipient's observed duration of graft function and using the pooled tail to account for the period after censoring. Under conditional independent censoring, Propositions~\ref{prop:target} and~\ref{prop:pooling} show that the population target recovers graft survival below $\tau$ for recipient profiles whose residual survival agrees with the pooled tail at the relevant censoring times. Theorems~\ref{thm:psi}--\ref{thm:argmin} establish that, as the cohort grows and optimization error vanishes, the base model approaches the best attainable population score for the pooled-completion target.
\end{remark}

\section{Simulation Experiments}\label{sec:simulations}
We conduct a range of simulation experiments to complement the established theory, as well as evaluate model performance across varied settings of sample sizes and censoring rates. Section \ref{sec:simulations_asymptotic} describes simulations related to the asymptotic results, while Section \ref{sec:simulations_performance} compares model performance on synthetic survival datasets. The simulation tables are deferred to supplementary Section~\ref{appendix:tables_figures} for brevity in the main text.

\begin{figure}[t]
    \centering
    \includegraphics[width=0.5\textwidth]{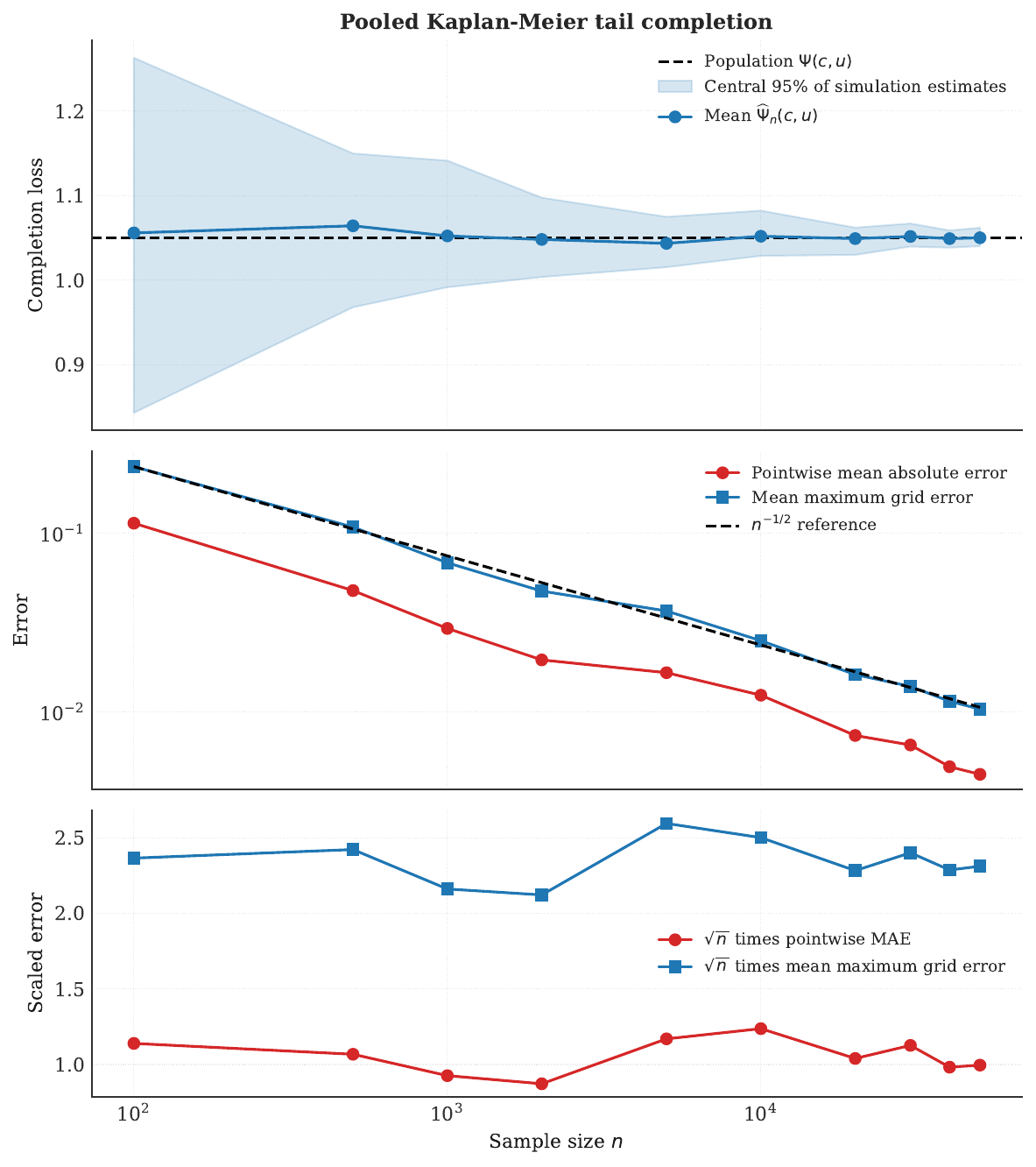}
    \caption{Empirical validation of Theorem~\ref{thm:psi}: pooled KM tail-completion estimation using
$50$ independent simulation repetitions and $B=50$ bootstrap
samples per dataset.
Top: mean estimate at $(c,u)=(e^{-1/2},0.2)$, its population
reference and the central empirical $95\%$ replicate interval;
Middle: pointwise mean absolute error and mean maximum error
over the fixed query grid, with an $n^{-1/2}$ reference;
Bottom: the corresponding $\sqrt n$-scaled errors.}
    \label{fig:sim_thm_1}
\end{figure}

\subsection{Simulations for Asymptotic Properties}\label{sec:simulations_asymptotic}
To empirically examine the convergence rate established in Theorem~\ref{thm:psi}, we conducted simulations across sample sizes ranging from $n=100$ to $50{,}000$, generating $50$ independent datasets per size. The covariates were drawn as $X\sim\mathcal N(0,I_5)$, with the true unrestricted log-time generated as $Z=\beta^\top X+\epsilon$, where $\beta=(1.5,-1.0,0.5,0.8,-0.2)^\top$ and $\epsilon\sim\mathcal N(0,1)$. The survival time is $T=e^Z$. Independent right-censoring was generated as $C=e^{-1}+\exp(\xi_C)$ with $\xi_C\sim\mathcal N(0,4)$, and the clinical horizon was fixed at $\tau=e^2$. Because $X$ and $\xi$ are zero-mean Gaussians, the marginal log-time is normally distributed with $Z \sim \mathcal{N}(0, \Vert{}\beta\Vert{}^2 + 1)$. This Gaussian marginalization permits an exact, analytical computation of the population target $\Psi(c,u)$. 
For each generated dataset, the estimated deviation $\widehat\Psi_n$ was computed by averaging $B=50$ bootstrap KM tails, placing any residual mass at $\tau$. Figure~\ref{fig:sim_thm_1} illustrates the estimator's behavior at a representative query point $(c,u)=(e^{-1/2},0.2)$, demonstrating that it robustly targets the exact analytical reference while its empirical replicate interval contracts as $n$ increases. Both the pointwise mean absolute error and the mean maximum error evaluated over a fixed, finite two-dimensional grid of $(c,u)$ values decay at the rate of $n^{-1/2}$. Scaling these grid errors by $\sqrt{n}$ yields approximately stable horizontal trajectories, providing empirical validation of the theoretical convergence rate derived in Theorem~\ref{thm:psi}.

\begin{figure}[t]
    \centering
    \includegraphics[width=0.9\textwidth]{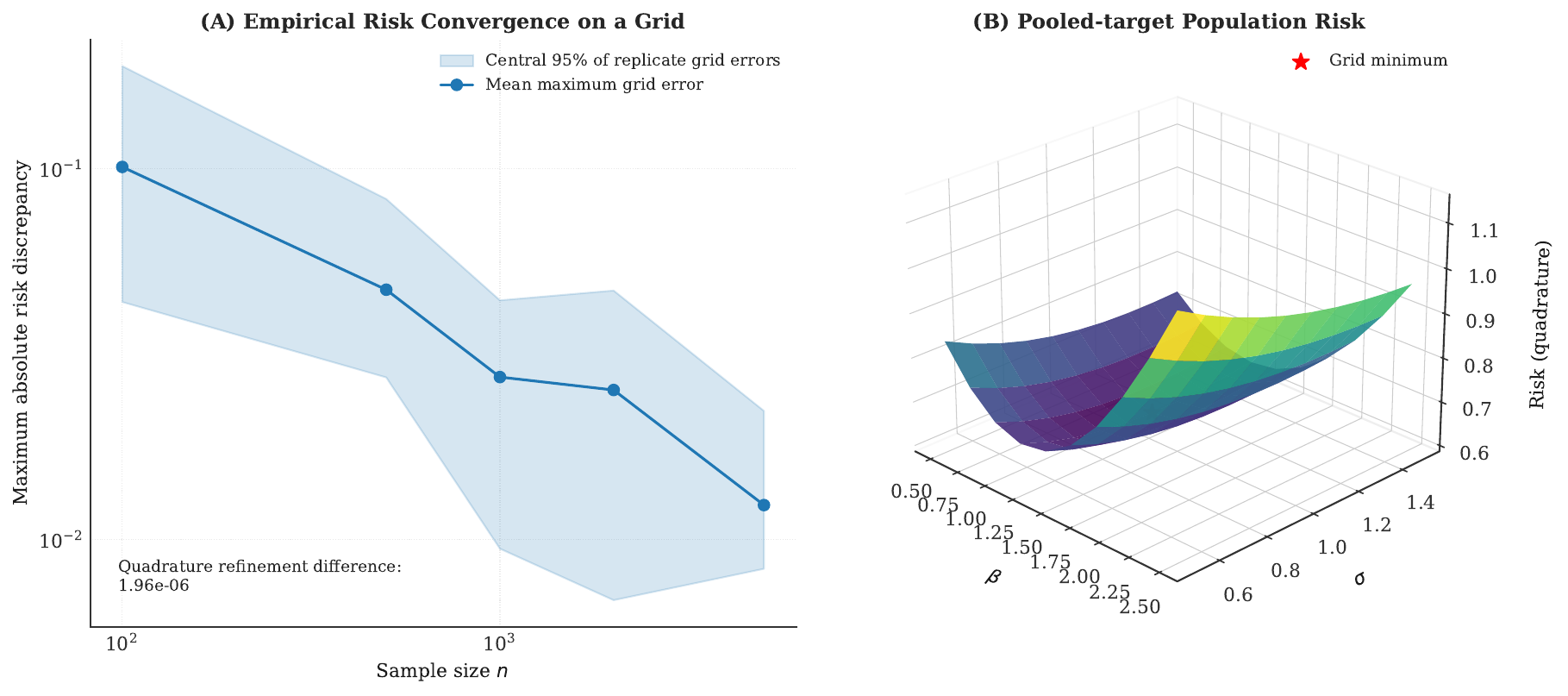}
    \caption{Finite-grid assessment of Theorem~\ref{thm:ulln}.
(A) Mean maximum absolute discrepancy between empirical and
quadrature-based population risks, with the central empirical
$95\%$ range across simulation repetitions.
(B) Pooled-target population risk over
$\Theta=[0.5,2.5]\times[0.5,1.5]$.}
    \label{fig:sim_thm_2}
\end{figure}

To examine Theorem~\ref{thm:ulln}, we used the family
$U_\nu(x,\epsilon)=\beta x+\sigma\epsilon$ on
$\Theta=[0.5,2.5]\times[0.5,1.5]$, where $\sigma$ denotes the noise scaler and $\epsilon \sim \mathcal N(0,1)$. Data were generated as $Z=1.5X+\epsilon$ and $T=e^Z$, with independent standard normal $X,\epsilon$, and independently $C=e^{-1}+\exp(\xi_C)$, where $\xi_C\sim\mathcal N(0,4)$. The horizon was fixed at $\tau=e^2$.
For each $n\in\{100,500,1000,2000,5000\}$, we generated $10$ independent datasets and evaluated $\widehat R_n$ on a
$10\times10$ parameter grid using $B=25$ bootstrap KM curves and $m=50$ observation-specific prediction noises, held fixed across parameter values. The pooled-target population risk was approximated by deterministic quadrature, with numerical stability assessed by successive refinement.
Figure~\ref{fig:sim_thm_2} shows the maximum absolute discrepancy between the empirical risk and the quadrature approximation to the pooled-target population risk over the parameter grid, averaged across replications
(panel~A), together with the population risk surface (panel~B). The mean maximum discrepancy decreases as the sample size increases, a pattern consistent with the uniform convergence in probability
established in Theorem~\ref{thm:ulln}.

\subsection{Simulations for Predictive Performance}\label{sec:simulations_performance}
\begin{figure}[!t]
    \centering
    \includegraphics[width=0.9\textwidth]{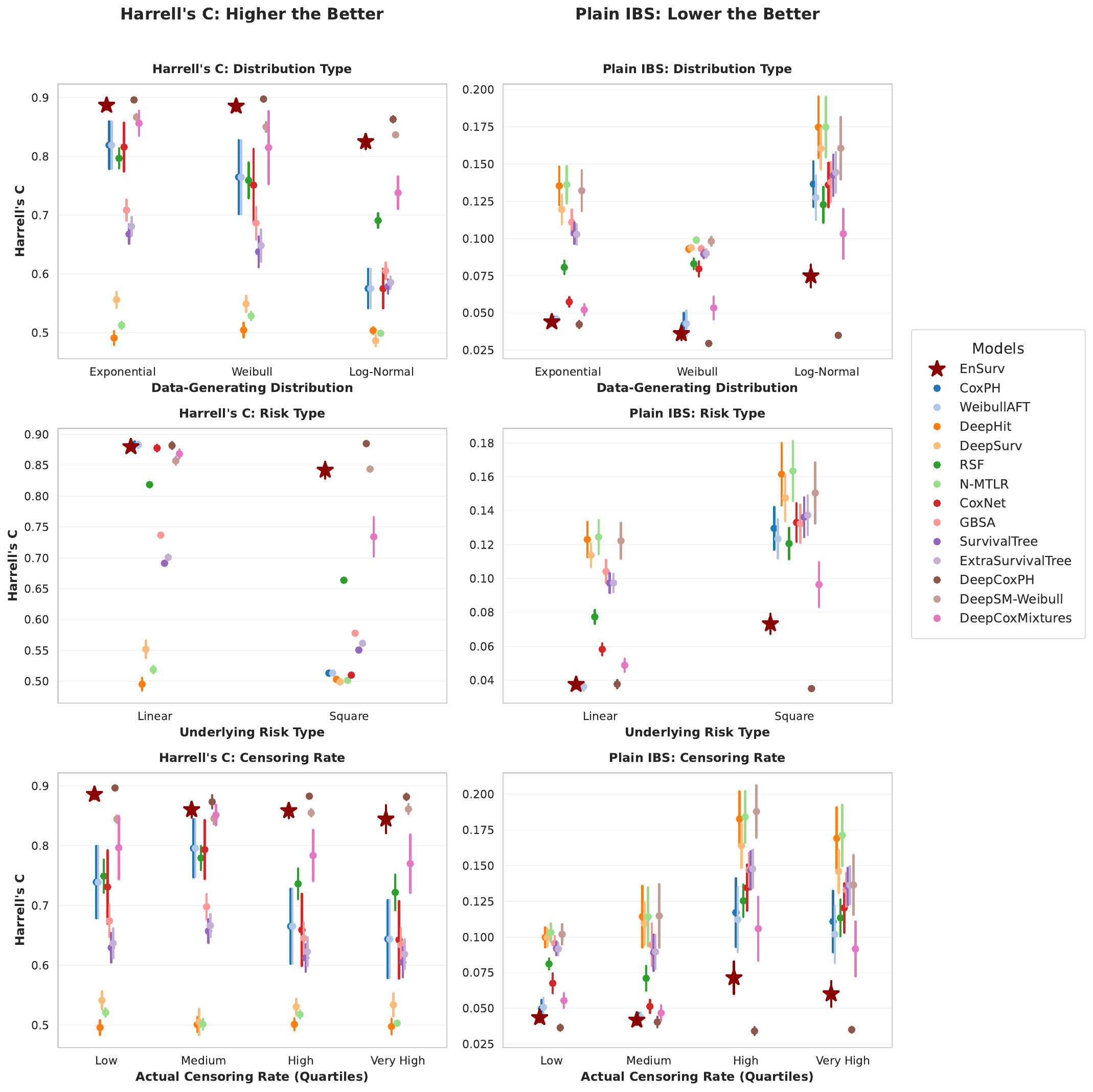}
    \caption{Simulation results for prediction with respect to Harrell's C (left panel) and Plain IBS (right panel) metrics, stratified by data-generating distribution (top row), risk type (middle row), and censoring rates (bottom row), averaged across all simulation settings in Table~\ref{table:simulation_results}.}
    \label{fig:pred_simulation_results}
\end{figure}
To evaluate the predictive performance of EnSurv against statistical and deep learning survival methods (see Table~\ref{table:model-comparison}), we conduct a comprehensive simulation study using synthetic survival datasets. 
The data generation process assumes three distinct baseline hazard distributions, namely, Exponential, Weibull, and Log-Normal, to capture a variety of temporal event dynamics and underlying survival trajectories. Moreover, we consider two structural forms for the true underlying risk function: linear and square configurations, to assess the model's capacity to capture both proportional and non-linear covariate effects. To simulate realistic right-censoring patterns, the censoring times are drawn using varying mean parameters of 5, 20, 35, and 50 from a normal distribution, which are controlled to maintain an overall censoring rate above 10\% across all datasets. These configurations are evaluated across multiple sample sizes ($n = 5000, 10000,$ and $20000$), resulting in a total of 39 distinct evaluation scenarios. Results of the simulation study detailed in Table~\ref{table:simulation_results} and visualized in Figure~\ref{fig:pred_simulation_results} demonstrate that EnSurv is a highly competitive framework for predicting time-to-event outcomes, consistently outperforming other methods across a range of evaluation metrics.

\section{Application to the Prediction of Kidney Graft Outcomes}\label{sec:real_data}
While the base form of our proposed framework is broadly applicable to any survival analysis domain assuming conditionally independent right censoring, this study focuses specifically on predicting kidney graft failure. To robustly model graft survival using the comprehensive, real-world UNOS registry, we integrate region-level covariates, which substantially impact transplant outcomes, alongside domain constraints on patient-specific clinical variables. Furthermore, EnSurv incorporates a linear component that directly elucidates feature importance, fulfilling a critical requirement for interpretability in medical predictive modeling. Throughout our experiments, we set $\tau$ to the maximum observed graft-failure time in the training cohort, thereby retaining every observed training failure at its recorded event time within the prediction window. Moreover, we partition our analysis to evaluate both pre- and post-transplant outcomes. Pre-transplant predictions facilitate a rigorous evaluation of kidney offers, whereas post-transplant forecasts empower clinicians to tailor personalized interventions aimed at prolonging graft survival. 

\subsection{Regional Effects and Domain Knowledge}\label{sec:domain}

To capture hierarchical region-level effects, we incorporated macro-structural covariates including geographic region, region volume, demographic skew, and relative regional donor market share. Beyond these, we engineered four contextual features to capture unobserved institutional phenotypes and risk profiles. Specifically, we derived a leave-one-out delayed graft function (DGF) rate to reflect systemic procurement or perioperative vulnerabilities; the Shannon entropy of region-specific immunosuppression regimens to proxy patient pool complexity and protocol customization; a region-level Patient Acuity Index (comprising regional mean donor risk and patient sensitization) to quantify regional tolerance for marginal organs; and an Operational Strain Ratio (median length of stay relative to region volume) to measure infrastructural fatigue. Detailed formulations for these contextual covariates are provided in supplementary Section~\ref{appendix:region_covariates}.

Next, to introduce domain constraints within the EnSurv framework, we impose a soft derivative penalty in the predicted log-time during backpropagation. For patient $i$ and covariate $k$, define the sample averaged derivative
\[
J_{ik}(\nu)=\frac1m\sum_{j=1}^m\frac{\partial u_{ij}(\nu)}{\partial x_{i,k}},
\]
where $u_{ij}(\nu)$ is the prediction of EnSurv under noise realization $\epsilon_{ij}$ (Section~\ref{sec:estimation}). Let $\mathcal C^+$ and $\mathcal C^-$ contain the covariates assigned positive and negative directions, respectively. The non-negative penalty and the corresponding regularized objective are respectively given by
\[
\mathcal P_n(\nu)=\frac1n\sum_{i=1}^n\left\{
\sum_{k\in\mathcal C^+}\max\left(0,-J_{ik}(\nu)\right)+
\sum_{k\in\mathcal C^-}\max\left(0,J_{ik}(\nu)\right)\right\},\qquad
\widehat R_n(\nu)+\lambda_n\mathcal P_n(\nu),
\]
where $\lambda_n$ denotes the constraint strength. The penalty encourages the model to explore clinically established directions with respect to covariates, while preserving its full capacity to discover complex interaction patterns among unconstrained covariates. 

\subsection{Experimental Setup}
To model post-transplant time to graft failure, our feature space incorporates 24 patient-level demographic and clinical covariates from the UNOS registry. As detailed in Table~\ref{table:data_explanation} (supplementary Section~\ref{appendix:tables_figures}), these variables include age, race, gender, body mass index (BMI), blood group, hepatitis C virus (HCV) status, donor risk index, creatinine levels at transplant and discharge, dialysis status, and immunosuppression strategies. We further augment these patient-specific features with the nine region-level covariates introduced in the preceding section. In contrast, predicting pre-transplant graft outcomes requires restricting the feature set to information available prior to the operation. Consequently, we exclude all post-operative covariates, namely first-week dialysis requirements, specific immunosuppression regimens, discharge creatinine levels, and post-transplant length of stay. Furthermore, in this pre-transplant configuration, the engineered regional covariates are limited exclusively to patient acuity indices.

Missing data were handled by applying median imputation for numerical covariates and a $20$-nearest neighbors approach for categorical variables. Nominal categorical variables such as blood group, region, and HCV status were one-hot encoded, leading to a collection of 48 features for model fitting. For stable gradient computation and updates, numeric covariates were standardized. For both pre- and post-transplant settings, the EnSurv model was implemented using a standard MLP architecture comprising two hidden layers of 512 dimensions each, supplemented with batch normalization. Latent noise was sampled from the $\mathrm{Unif}(-1,1)$ distribution. For backpropagation, we employed the Adam optimizer with a learning rate of $10^{-3}$, training the model for 50 epochs with an early-stopping patience of 10 epochs. To mitigate bias arising from random data partitioning, the evaluation was conducted via a stratified 3-fold cross-validation framework that preserved the censoring rate across all folds. 

Domain constraints were imposed such that higher KDRI and serum creatinine levels, and the requirement for first-week dialysis are learned to be associated with adverse effects on post-transplant graft survival, while donor–recipient HCV negativity was specified to confer a protective benefit. Model performance was assessed using a comprehensive suite of metrics following \cite{qi2023survivaleval} for evaluating all three aspects: discrimination, accuracy, and calibration, including Harrell's, Antolini's, and Uno's concordance indices (C-indices), hinge and margin errors, uncensored mean absolute error (MAE), integrated Brier score (IBS), area under the precision-recall curve (AUPRC), and Kaplan-Meier (KM) calibration. Time-dependent metrics such as Brier scores, area under the ROC curve (AUC), and Uno's C-indices were reported at 1- and 5-year horizons on a grid of 100 time points. A brief description of the metrics is provided in supplementary Section~\ref{appendix:metrics}. Finally, EnSurv was benchmarked against a selection of competitive baseline survival models (Table~\ref{table:model-comparison}), including statistical (CoxPH, WeibullAFT, CoxNet), machine learning (SurvivalTree, ExtraSurvivalTree, RSF, GBSA), deep learning (DeepSurv, DeepHit, N-MTLR, DeepCoxPH, DeepCoxMixtures, DeepSurvivalMachines), and foundation model (SurvivalPFN). All models were trained using the identical covariate set, data splits, and hyperparameter configurations to ensure a fair evaluation, following \cite{wang2026survival}. The following sections detail the experimental results for the post-transplantation setting; quantitative pre-transplantation results are provided in supplementary Section~\ref{appendix:tables_figures}.

\begin{table}[t]
\centering
\caption{Comparison of EnSurv against survival analysis baselines across key modeling characteristics.}
\label{table:model-comparison}
\resizebox{\textwidth}{!}{%
\begin{adjustbox}{width=\textwidth}
\begin{tabular}{l*{9}{c}}
\toprule
\textbf{Model} &
\makecell{Deep\\(NN-based)} &
\makecell{Nonlinear\\covariate effects} &
\makecell{Distribution-\\free} &
\makecell{No PH\\assumption} &
\makecell{No time\\discretization} &
\makecell{No separate baseline-\\hazard estimation} &
\makecell{Explainability} &
\makecell{Domain-\\knowledge} & Reference\\
\midrule

CoxPH & \xmark & \xmark & \cmark & \xmark & \cmark & \xmark & \cmark & \xmark & \cite{cox1972regression} \\ 
        WeibullAFT & \xmark & \xmark & \xmark & \xmark & \cmark & \cmark & \cmark & \xmark & \cite{wei1992accelerated} \\ 
        DeepHit & \cmark & \cmark & \cmark & \cmark & \xmark & \cmark & \xmark & \xmark & \cite{lee2018deephit} \\ 
        DeepSurv & \cmark & \cmark & \cmark & \xmark & \cmark & \xmark & \xmark & \xmark & \cite{katzman2018deepsurv} \\ 
        Random Survival Forests (RSF) & \xmark & \cmark & \cmark & \cmark & \cmark & \cmark & \xmark & \xmark & \cite{ishwaran2008random} \\ 
        Neural Multi-Task Logistic Regression (N-MTLR) & \cmark & \cmark & \cmark & \cmark & \xmark & \cmark & \xmark & \xmark & \cite{fotso2018deep} \\ 
        CoxNet & \xmark & \xmark & \cmark & \xmark & \cmark & \xmark & \cmark & \xmark & \cite{simon2011regularization} \\ 
        GBSA & \xmark & \cmark & \cmark & \xmark & \cmark & \xmark & \xmark & \xmark & \cite{hothorn2006survival} \\ 
        SurvivalTree & \xmark & \cmark & \cmark & \cmark & \cmark & \cmark & \xmark & \xmark & \cite{leblanc1993survival} \\ 
        ExtraSurvivalTree & \xmark & \cmark & \cmark & \cmark & \cmark & \cmark & \xmark & \xmark & \cite{sksurv} \\ 
        DeepCoxPH & \cmark & \cmark & \cmark & \xmark & \cmark & \xmark & \xmark & \xmark & \cite{faraggi1995neural} \\ 
        DeepSurvivalMachines (DSM) & \cmark & \cmark & \xmark & \cmark & \cmark & \cmark & \xmark & \xmark & \cite{nagpal2021dsm} \\ 
        DeepCoxMixtures (DCM) & \cmark & \cmark & \cmark & \cmark & \cmark & \xmark & \xmark & \xmark & \cite{nagpal2021dcm} \\ 
        Survival Prior-data Fitted Network (SurvivalPFN) & \cmark & \cmark & \cmark & \cmark & \xmark & \cmark & \xmark & \xmark & \cite{qi2026survivalpfn} \\ 
        \textbf{EnSurv} & \cmark & \cmark & \cmark & \cmark & \cmark & \cmark & \cmark & \cmark & (Proposed) \\ 
\bottomrule
\end{tabular}%
\end{adjustbox}
}
\end{table}

\subsection{Experimental Results}
\begin{figure}[t]
\centering
\includegraphics[width=\textwidth]{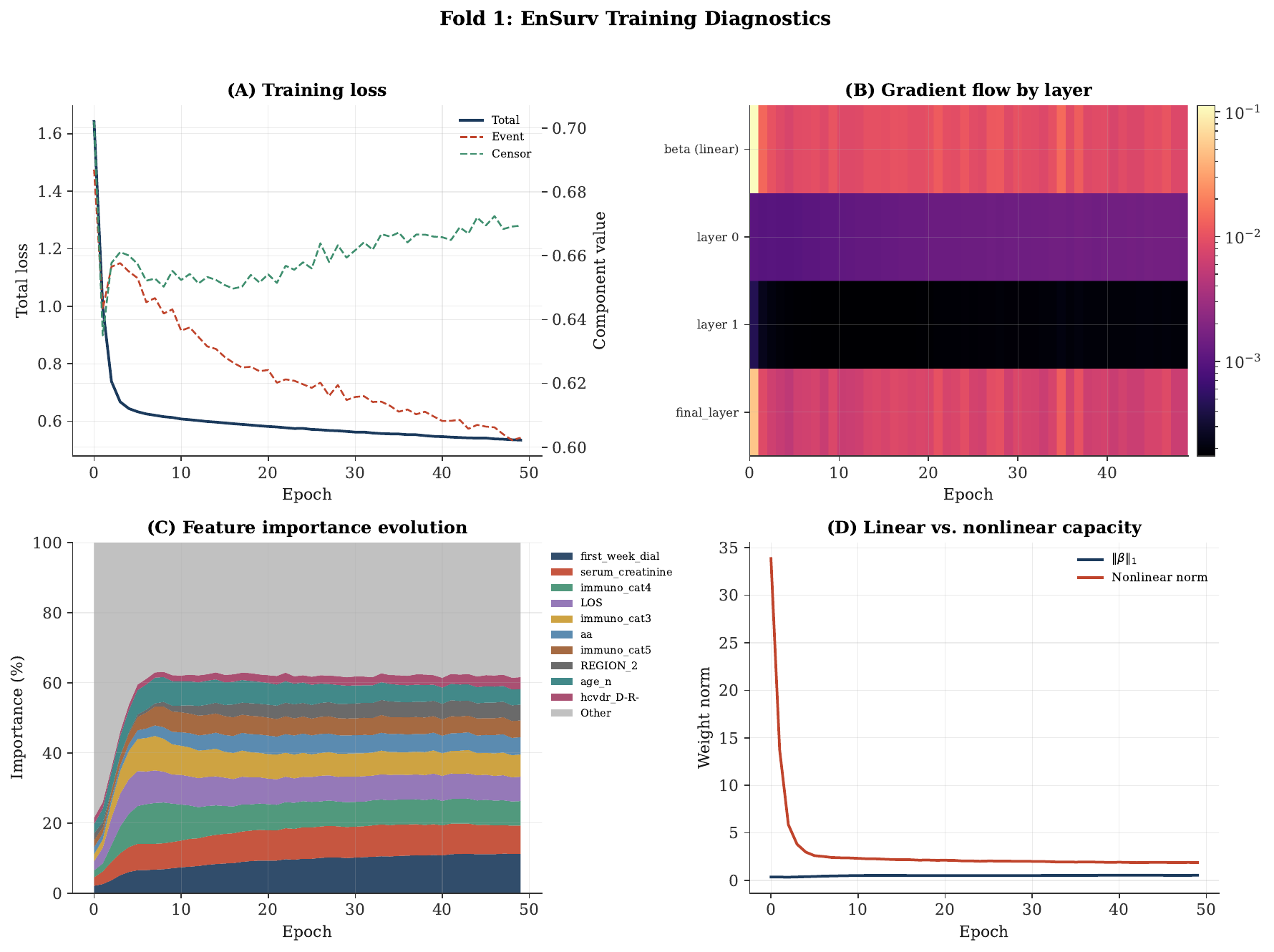}
\caption{Training diagnostics of EnSurv on fold 1 of the UNOS dataset in the post-transplantation setting.}
\label{fig:training_diagnostics}
\end{figure}

The training diagnostics presented in Figure~\ref{fig:training_diagnostics} detail EnSurv’s optimization dynamics and feature importance evolution during training on the first fold for the post-transplant scenario. Panel (A) illustrates empirical convergence over $50$ epochs. Panel (B) reveals the gradient flow: the linear $\beta$ parameters and final output layer sustain robust norms ($\sim$$10^{-2}$), whereas deeper non-linear representations (layer 1) attenuate significantly ($\sim$$10^{-4}$). This mirrors the capacity allocation in Panel (D), where the non-linear weight norm drops from $\sim 35$ to $<5$ within five epochs to prevent overfitting. 
Panel (C) shows stacked feature-importance trajectory evolution during training. The model isolates core clinical drivers, notably first week dialysis requirement, recipient's creatinine level at discharge, and specific demographic characteristics (e.g., age, race), which consistently account for $\sim$60\% of the total predictive importance. This demonstrates that EnSurv successfully leverages its non-linear transformations while preserving clinical interpretability.

\begin{figure}[t]
    \centering
    \includegraphics[width=0.8\textwidth]{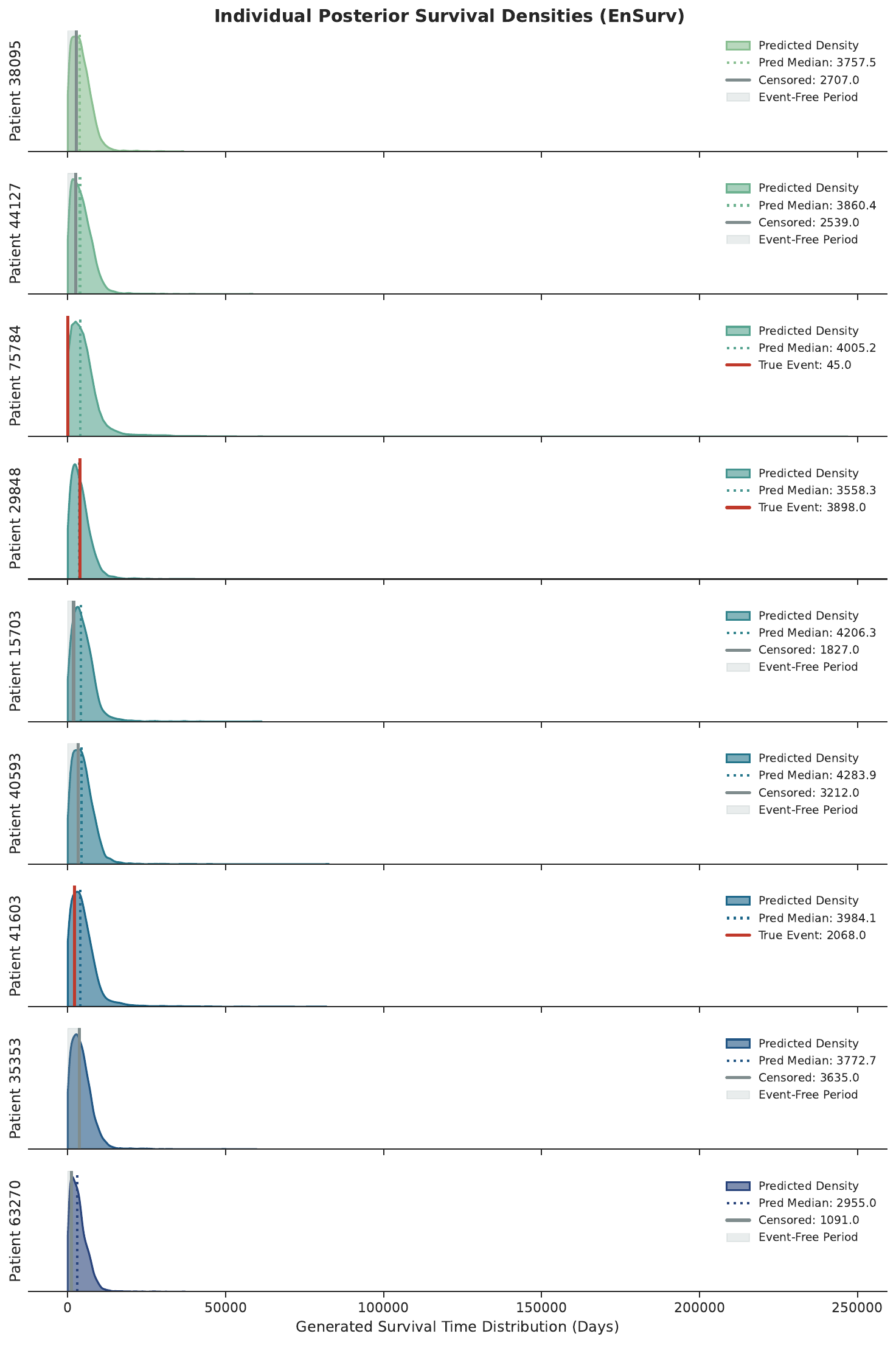}
    \caption{EnSurv-predicted survival distributions for random patients in the test set of fold 1.}
    \label{fig:individual_survival}
\end{figure}

\begin{figure}[t]
    \centering
    \includegraphics[width=0.9\textwidth]{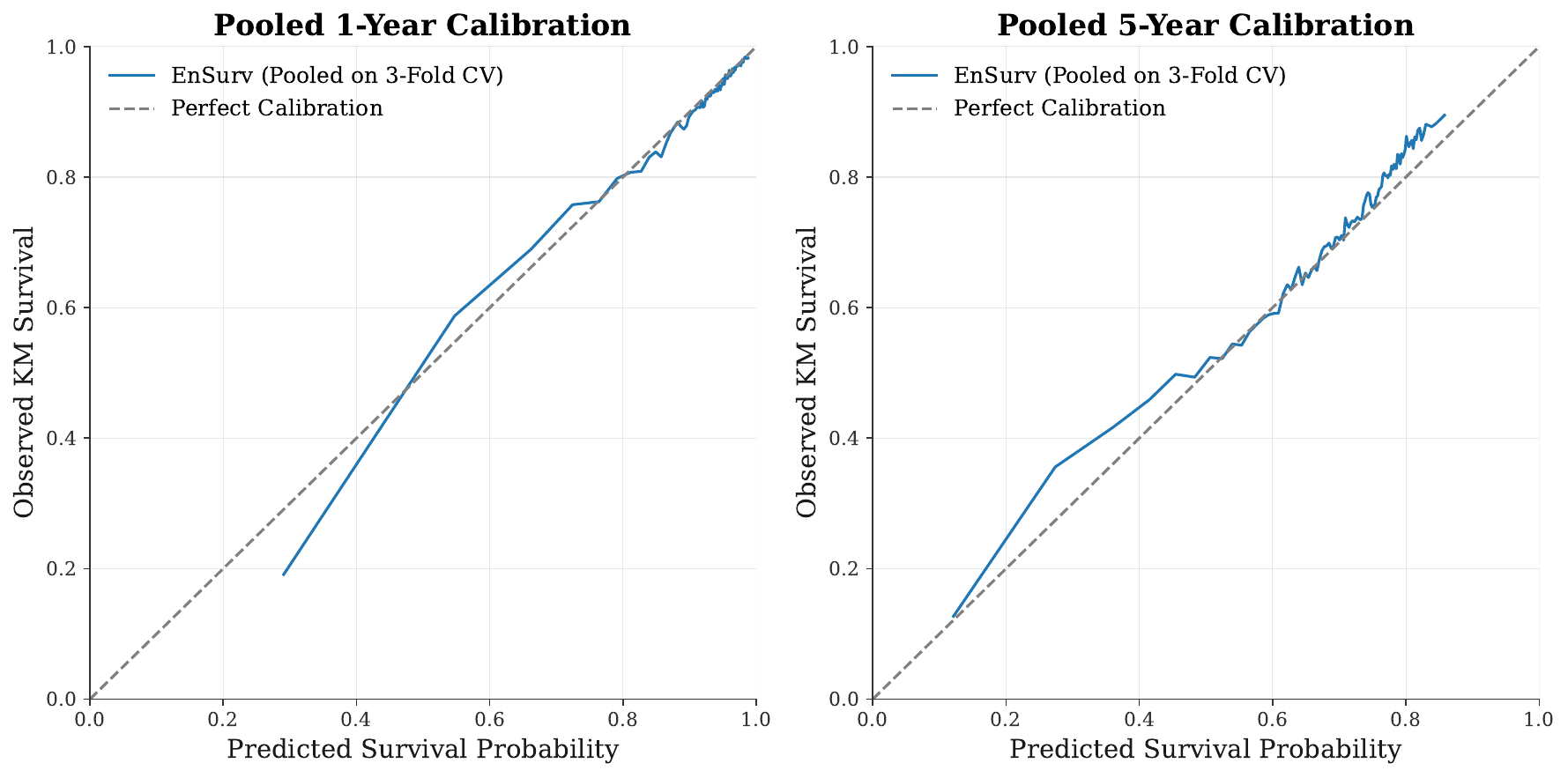}
    \caption{1- and 5-year pooled calibration plots for the EnSurv model on the post-transplant setting. Out-of-fold predicted survival probabilities from 3-fold cross-validation are stratified into quantiles and plotted against corresponding empirical KM estimates.}
    \label{fig:pooled_calibration}
\end{figure}

Table~\ref{table:kidney_results} reports the quantitative results obtained by EnSurv and the baseline models on each fold of the UNOS dataset in the post-transplant setting, along with the cross-validation mean and standard deviation. EnSurv consistently outperforms its competitors; yielding the highest concordance indices highlights its superior ability to rank graft failure events, while lower IBS scores signify accurate probabilistic forecasting. 
Figure~\ref{fig:individual_survival} plots the survival distributions of random patients from the test set of the first fold, obtained by drawing 5,000 samples from the trained EnSurv model. Finally, to visualize calibration at clinically established post-transplantation milestones \citep{poggio2021long}, out-of-fold predictions from all three cross-validation partitions were pooled to construct 1- and 5-year calibration plots (Figure~\ref{fig:pooled_calibration}). The close adherence of EnSurv's predicted probabilities to the empirical KM estimates alongside the reference diagonal indicates a well-calibrated model.

Consistent with these post-transplant findings, our pre-transplant modeling demonstrates similar optimization stability and predictive superiority. Despite being restricted entirely to pre-operative covariates, EnSurv maintains its competitive advantage over the baseline suite, yielding robust concordance indices and well-calibrated survival estimates for kidney offer evaluations. To maintain narrative focus, the corresponding quantitative metrics for the pre-transplant scenario are provided in supplementary Section~\ref{appendix:tables_figures}.

\begin{table}[!t]
\caption{Results for modeling the time to graft failure on 3 folds of the UNOS dataset under the post-transplant setting. The \underline{\textbf{best}} and \textbf{second-best} results are highlighted. We note that because KM estimators at the terminal nodes of tree-based survival models (SurvivalTree and ExtraSurvivalTree) can plateau above a probability of 0.5 under heavy right-censoring, their median survival times frequently become undefined. Consequently, point-based error metrics yield infinite values for these specific models and are omitted from our results to maintain a consistent comparative framework across all evaluated architectures.} \label{table:kidney_results}
\begin{adjustbox}{width=\textwidth}
\begin{tabular}{cccccccccccccccccc}
\toprule
Fold & Model & \makecell{Harrell's\\C-index} & \makecell{Antolini's\\C-index} & \makecell{Uno's\\C-index} & \makecell{Hinge\\Error} & \makecell{Margin\\Error} & \makecell{Uncensored\\MAE} & \makecell{Plain\\IBS} & \makecell{IBS-\\IPCW} & AUPRC & \makecell{KM-Calib-\\ration} & \makecell{AUC\\1-Yr} & \makecell{Brier\\1-Yr} & \makecell{AUC\\5-Yr} & \makecell{Brier\\5-Yr} & \makecell{Uno's C\\1-Yr} & \makecell{Uno's C\\5-Yr} \\ \midrule
        1 & CoxPH & 0.6453 & 0.6453 & 0.6133 & 931.4438 & 2090.8623 & 1981.8412 & 0.0696 & 0.1627 & $\mathbf{0.6609}$ & $\mathbf{0.0001}$ & 0.7273 & 0.0708 & 0.6659 & 0.186 & 0.7162 & 0.643 \\ 
        ~ & WeibullAFT & 0.6457 & 0.6457 & 0.6104 & 983.568 & 2177.3881 & 2107.5404 & 0.0755 & 0.1703 & 0.6479 & 0.0047 & 0.7315 & 0.0706 & 0.6618 & 0.1884 & 0.7214 & 0.644 \\ 
        ~ & CoxNet & 0.6452 & 0.6452 & 0.613 & 1928.09 & 3201.426 & 1938.7674 & 0.1738 & 0.3559 & 0.1465 & 0.1935 & 0.728 & 0.79 & 0.6659 & 0.6168 & 0.7168 & 0.6427 \\ 
        ~ & RSF & 0.6357 & 0.6418 & 0.6055 & 1926.9375 & 3200.2849 & 1937.4768 & 0.1677 & 0.3456 & 0.1583 & 0.1772 & 0.6657 & 0.753 & 0.6221 & 0.5938 & 0.7004 & 0.6322 \\ 
        ~ & GBSA & 0.6443 & 0.6443 & 0.6141 & 1929.468 & 3202.4643 & 1939.2349 & 0.1824 & 0.375 & 0.1281 & 0.2082 & 0.7167 & 0.8282 & 0.6603 & 0.651 & 0.7101 & 0.6427 \\ 
        ~ & SurvivalTree & 0.5944 & 0.6011 & 0.5776 & - & - & - & 0.169 & 0.3455 & 0.1595 & 0.1777 & 0.6079 & 0.7586 & 0.5941 & 0.5937 & 0.635 & 0.5921 \\ 
        ~ & ExtraSurvivalTree & 0.5814 & 0.5874 & 0.5674 & - & - & - & 0.1743 & 0.356 & 0.1468 & 0.1892 & 0.5916 & 0.7861 & 0.5801 & 0.6152 & 0.6205 & 0.5782 \\ 
        ~ & DeepHit & 0.5287 & 0.4975 & 0.5249 & 1931.8057 & 3204.7424 & 1941.3887 & 0.1978 & 0.4071 & 0.1028 & 0.2405 & 0.4732 & 0.9017 & 0.4917 & 0.7084 & 0.5345 & 0.53 \\ 
        ~ & DeepSurv & 0.5275 & 0.5275 & 0.5183 & 1919.4574 & 3192.5886 & $\mathbf{1929.9947}$ & 0.1404 & 0.2831 & 0.223 & 0.1148 & 0.551 & 0.5776 & 0.5403 & 0.4593 & 0.5464 & 0.5259 \\ 
        ~ & NMTLR & 0.5017 & 0.5032 & 0.5012 & 1931.6629 & 3204.616 & 1941.2666 & 0.2002 & 0.4121 & 0.1002 & 0.2456 & 0.5084 & 0.9129 & 0.5032 & 0.5067 & 0.7172 & 0.5004 \\ 
        ~ & DeepCoxPH & 0.6452 & 0.6452 & 0.6139 & 927.0772 & 2086.5859 & 1968.3674 & 0.0693 & 0.1617 & 0.6606 & \underline{$\mathbf{0.0000}$} & 0.7257 & 0.0709 & 0.6663 & 0.1861 & 0.7145 & 0.6427 \\ 
        ~ & DSM-Weibull & 0.4713 & 0.4807 & 0.4876 & 1931.6013 & 3204.5515 & 1941.2053 & 0.2014 & 0.4144 & 0.099 & 0.248 & 0.4292 & 0.9181 & 0.4844 & 0.7213 & 0.4267 & 0.475 \\ 
        ~ & DSM-LogNormal & 0.4614 & 0.4461 & 0.4982 & 1931.5455 & 3204.5491 & 1941.2483 & 0.2012 & 0.4142 & 0.0994 & 0.2474 & 0.3406 & 0.9169 & 0.4301 & 0.721 & 0.3607 & 0.4679 \\ 
        ~ & DeepCoxMixtures & $\mathbf{0.6485}$ & $\mathbf{0.6521}$ & $\mathbf{0.6157}$ & $\mathbf{918.5169}$ & $\mathbf{2072.0527}$ & 1941.7903 & $\mathbf{0.0683}$ & $\mathbf{0.1614}$ & \underline{$\mathbf{0.6631}$} & 0.0006 & $\mathbf{0.7439}$ & $\mathbf{0.0680}$ & $\mathbf{0.6680}$ & $\mathbf{0.1852}$ & $\mathbf{0.7231}$ & $\mathbf{0.6453}$ \\ 
        ~ & SurvivalPFN & 0.6301 & 0.631 & 0.5957 & 940.5572 & 2155.8453 & 1965.893 & 0.0696 & 0.167 & 0.659 & 0.0013 & 0.7197 & 0.0724 & 0.6544 & 0.1922 & 0.7029 & 0.6289 \\ 
        ~ & \textbf{EnSurv} & \underline{$\mathbf{0.6625}$} & \underline{$\mathbf{0.6595}$} & \underline{$\mathbf{0.6218}$} & \underline{$\mathbf{854.9591}$} & \underline{$\mathbf{2003.2349}$} & \underline{$\mathbf{1684.5283}$} & \underline{$\mathbf{0.0669}$} & \underline{$\mathbf{0.1607}$} & 0.6573 & 0.0006 & \underline{$\mathbf{0.7588}$} & \underline{$\mathbf{0.0648}$} & \underline{$\mathbf{0.6786}$} & \underline{$\mathbf{0.1848}$} & \underline{$\mathbf{0.7504}$} & \underline{$\mathbf{0.6604}$} \\ \midrule
        2 & CoxPH & 0.6461 & 0.6461 & 0.6146 & 916.5474 & 2074.4966 & 1949.4476 & 0.0688 & 0.1629 & 0.6613 & \underline{$\mathbf{0.0002}$} & 0.726 & 0.0685 & 0.6633 & 0.1857 & 0.7159 & 0.6422 \\ 
        ~ & WeibullAFT & 0.6468 & 0.6468 & 0.6119 & 967.4373 & 2157.9339 & 2071.7427 & 0.0748 & 0.1707 & 0.6479 & 0.0053 & 0.7301 & 0.0684 & 0.6596 & 0.189 & 0.7209 & 0.6436 \\ 
        ~ & CoxNet & 0.6457 & 0.6457 & 0.6144 & 1939.7399 & 3231.4408 & 1968.5833 & 0.1752 & 0.3588 & 0.1465 & 0.1958 & 0.7262 & 0.7929 & 0.663 & 0.6267 & 0.7159 & 0.6418 \\ 
        ~ & RSF & 0.6436 & 0.648 & 0.6115 & 1938.3312 & 3230.0466 & 1967.1374 & 0.1665 & 0.3429 & 0.1621 & 0.1756 & 0.6921 & 0.7445 & 0.6391 & 0.5927 & 0.7143 & 0.6402 \\ 
        ~ & GBSA & 0.6425 & 0.6425 & 0.6138 & 1941.0691 & 3232.4514 & 1969.0877 & 0.1825 & 0.3752 & 0.1298 & 0.2081 & 0.7108 & 0.8245 & 0.6565 & 0.6556 & 0.7046 & 0.64 \\ 
        ~ & SurvivalTree & 0.5949 & 0.5974 & 0.5735 & - & - & - & 0.1711 & 0.3517 & 0.1622 & 0.1773 & 0.6153 & 0.7575 & 0.5832 & 0.6051 & 0.6455 & 0.5919 \\ 
        ~ & ExtraSurvivalTree & 0.5972 & 0.6009 & 0.5809 & 1947.5705 & 3259.7816 & 2000.2169 & 0.1681 & 0.3422 & 0.1616 & 0.1769 & 0.6198 & 0.7515 & 0.5969 & 0.595 & 0.6415 & 0.5936 \\ 
        ~ & DeepHit & 0.4578 & 0.4932 & 0.4671 & 1943.5087 & 3234.8313 & 1971.3279 & 0.1992 & 0.4098 & 0.1028 & 0.2428 & 0.5217 & 0.9037 & 0.5274 & 0.718 & 0.4342 & 0.4607 \\ 
        ~ & DeepSurv & 0.5095 & 0.5095 & 0.5134 & 167123.1032 & 292424.5759 & 413870.9518 & 0.1781 & 0.3658 & 0.1399 & 0.1959 & 0.5154 & 0.7985 & 0.5061 & 0.6363 & 0.5156 & 0.5064 \\ 
        ~ & NMTLR & 0.519 & 0.5184 & 0.5057 & 1940.7265 & 3232.0448 & 1968.6557 & 0.1801 & 0.3701 & 0.1339 & 0.2029 & 0.558 & 0.8118 & 0.5442 & 0.5339 & 0.6445 & 0.518 \\ 
        ~ & DeepCoxPH & 0.646 & 0.646 & 0.6145 & 916.3224 & 2075.0483 & 1948.1318 & 0.0688 & 0.1629 & 0.6613 & \underline{$\mathbf{0.0002}$} & 0.7261 & 0.0686 & 0.663 & 0.1857 & 0.7158 & 0.6421 \\ 
        ~ & DSM-LogNormal & 0.4606 & 0.4469 & 0.498 & 1943.2806 & 3234.6738 & 1971.2359 & 0.2028 & 0.4173 & 0.0991 & 0.25 & 0.3417 & 0.9196 & 0.4258 & 0.7315 & 0.3616 & 0.4656 \\ 
        ~ & DeepCoxMixtures & $\mathbf{0.6485}$ & $\mathbf{0.6520}$ & $\mathbf{0.6162}$ & $\mathbf{904.3025}$ & $\mathbf{2056.6017}$ & $\mathbf{1910.1546}$ & $\mathbf{0.0676}$ & $\mathbf{0.1617}$ & $\mathbf{0.6637}$ & $\mathbf{0.0005}$ & $\mathbf{0.7400}$ & $\mathbf{0.0660}$ & $\mathbf{0.6654}$ & $\mathbf{0.1850}$ & $\mathbf{0.7217}$ & $\mathbf{0.6440}$ \\ 
        ~ & SurvivalPFN & 0.6316 & 0.6325 & 0.599 & 925.1714 & 2141.645 & 1936.6421 & 0.0689 & 0.1669 & 0.6597 & 0.001 & 0.7147 & 0.07 & 0.6532 & 0.1912 & 0.6991 & 0.6286 \\ 
        ~ & \textbf{EnSurv} & \underline{$\mathbf{0.6618}$} & \underline{$\mathbf{0.6574}$} & \underline{$\mathbf{0.6224}$} & \underline{$\mathbf{841.9854}$} & \underline{$\mathbf{1999.9217}$} & \underline{$\mathbf{1647.4140}$} & \underline{$\mathbf{0.0658}$} & \underline{$\mathbf{0.1606}$} & \underline{$\mathbf{0.6649}$} & 0.0007 & \underline{$\mathbf{0.7506}$} & \underline{$\mathbf{0.0632}$} & \underline{$\mathbf{0.6776}$} & \underline{$\mathbf{0.1842}$} & \underline{$\mathbf{0.7435}$} & \underline{$\mathbf{0.6594}$} \\ \midrule
        3 & CoxPH & 0.6469 & 0.6469 & 0.6153 & 918.0735 & 2071.9362 & 1954.7677 & 0.069 & 0.1605 & 0.6615 & \underline{$\mathbf{0.0001}$} & 0.7274 & 0.0701 & 0.6642 & 0.185 & 0.7171 & 0.6427 \\ 
        ~ & WeibullAFT & 0.648 & 0.648 & 0.6129 & 966.562 & 2148.9937 & 2072.6449 & 0.0749 & 0.1685 & 0.648 & 0.005 & 0.731 & 0.07 & 0.6611 & 0.1879 & 0.7217 & $\mathbf{0.6447}$ \\ 
        ~ & CoxNet & 0.6471 & 0.6471 & 0.6154 & 1932.3043 & 3214.4126 & 1954.4558 & 0.175 & 0.3559 & 0.1461 & 0.1986 & 0.7282 & 0.794 & 0.6646 & 0.6229 & 0.7177 & 0.6429 \\ 
        ~ & RSF & 0.6429 & 0.6481 & 0.6128 & 1931.2085 & 3213.3675 & 1953.3331 & 0.1683 & 0.3442 & 0.1582 & 0.1817 & 0.6785 & 0.7554 & 0.6324 & 0.5976 & 0.7089 & 0.6398 \\ 
        ~ & GBSA & 0.6418 & 0.6418 & 0.6135 & 1933.711 & 3215.5414 & 1955.0819 & 0.1833 & 0.3743 & 0.1281 & 0.2129 & 0.707 & 0.8306 & 0.6574 & 0.6556 & 0.7001 & 0.639 \\ 
        ~ & SurvivalTree & 0.6096 & 0.6119 & 0.5862 & - & - & - & 0.1666 & 0.3406 & 0.1693 & 0.1726 & 0.6334 & 0.7388 & 0.6042 & 0.5846 & 0.6602 & 0.6081 \\ 
        ~ & ExtraSurvivalTree & 0.5971 & 0.6018 & 0.5769 & - & - & - & 0.171 & 0.3477 & 0.155 & 0.1856 & 0.6106 & 0.7689 & 0.5964 & 0.6036 & 0.6414 & 0.5943 \\ 
        ~ & DeepHit & 0.4821 & 0.4915 & 0.4916 & 1936.0487 & 3217.8272 & 1957.2413 & 0.1985 & 0.4055 & 0.1034 & 0.2449 & 0.4922 & 0.9025 & 0.5054 & 0.7119 & 0.4689 & 0.4824 \\ 
        ~ & DeepSurv & 0.4395 & 0.4395 & 0.4608 & 2045.0601 & 3417.5123 & 2237.5576 & 0.1794 & 0.3666 & 0.1389 & 0.1968 & 0.3856 & 0.8002 & 0.4266 & 0.6365 & 0.3923 & 0.4417 \\ 
        ~ & NMTLR & 0.4976 & 0.4972 & 0.4965 & 1931.714 & 3219.9471 & 1964.0556 & 0.1691 & 0.345 & 0.1569 & 0.1785 & 0.4993 & 0.7512 & 0.4963 & 0.5053 & 0.5937 & 0.4986 \\ 
        ~ & DeepCoxPH & 0.6468 & 0.6468 & 0.6157 & 915.0955 & 2069.535 & 1946.8987 & 0.0687 & 0.1596 & 0.6613 & \underline{$\mathbf{0.0001}$} & 0.7268 & 0.0702 & 0.6647 & 0.185 & 0.7163 & 0.6425 \\ 
        ~ & DSM-LogNormal & 0.4569 & 0.445 & 0.4905 & 1935.6474 & 3217.5168 & 1957.0049 & 0.2019 & 0.4128 & 0.1004 & 0.2518 & 0.3402 & 0.9173 & 0.4227 & 0.725 & 0.3609 & 0.4642 \\ 
        ~ & DeepCoxMixtures & $\mathbf{0.6495}$ & $\mathbf{0.6526}$ & $\mathbf{0.6172}$ & $\mathbf{906.6706}$ & $\mathbf{2055.8258}$ & $\mathbf{1920.8375}$ & \underline{$\mathbf{0.0678}$} & $\mathbf{0.1589}$ & $\mathbf{0.6639}$ & $\mathbf{0.0005}$ & $\mathbf{0.7413}$ & $\mathbf{0.0674}$ & $\mathbf{0.6661}$ & $\mathbf{0.1843}$ & $\mathbf{0.7230}$ & 0.6446 \\ 
        ~ & SurvivalPFN & 0.6328 & 0.6328 & 0.6006 & 930.3229 & 2144.3101 & 1949.166 & 0.0692 & 0.1644 & 0.6595 & 0.0011 & 0.7159 & 0.0718 & 0.653 & 0.191 & 0.701 & 0.629 \\ 
        ~ & \textbf{EnSurv} & \underline{$\mathbf{0.6633}$} & \underline{$\mathbf{0.6612}$} & \underline{$\mathbf{0.6244}$} & \underline{$\mathbf{867.4175}$} & \underline{$\mathbf{2003.2319}$} & \underline{$\mathbf{1786.8035}$} & $\mathbf{0.0682}$ & \underline{$\mathbf{0.1575}$} & \underline{$\mathbf{0.6693}$} & 0.0006 & \underline{$\mathbf{0.7535}$} & \underline{$\mathbf{0.0646}$} & \underline{$\mathbf{0.6783}$} & \underline{$\mathbf{0.1827}$} & \underline{$\mathbf{0.7437}$} & \underline{$\mathbf{0.6599}$} \\ \midrule
        Mean $\pm$ SD & CoxPH & $\xms{0.6461}{0.0008}$ & $\xms{0.6461}{0.0008}$ & $\xms{0.6144}{0.0010}$ & $\xms{922.0216}{8.1955}$ & $\xms{2079.0984}{10.2680}$ & $\xms{1962.0188}{17.3716}$ & $\xms{0.0691}{0.0004}$ & $\xms{0.1620}{0.0013}$ & $\xms{0.6612}{0.0003}$ & $\xubms{0.0001}{0.0001}$ & $\xms{0.7269}{0.0008}$ & $\xms{0.0698}{0.0011}$ & $\xms{0.6645}{0.0013}$ & $\xms{0.1856}{0.0005}$ & $\xms{0.7164}{0.0005}$ & $\xms{0.6427}{0.0003}$ \\ 
        ~ & WeibullAFT & $\xms{0.6469}{0.0011}$ & $\xms{0.6469}{0.0011}$ & $\xms{0.6117}{0.0013}$ & $\xms{972.5225}{9.5757}$ & $\xms{2161.4386}{14.5180}$ & $\xms{2083.9760}{20.4124}$ & $\xms{0.0751}{0.0004}$ & $\xms{0.1698}{0.0012}$ & $\xms{0.6479}{0.0001}$ & $\xms{0.0050}{0.0003}$ & $\xms{0.7308}{0.0007}$ & $\xms{0.0697}{0.0011}$ & $\xms{0.6608}{0.0011}$ & $\xms{0.1884}{0.0006}$ & $\xms{0.7214}{0.0003}$ & $\xms{0.6441}{0.0005}$ \\ 
        ~ & CoxNet & $\xms{0.6460}{0.0010}$ & $\xms{0.6460}{0.0010}$ & $\xms{0.6143}{0.0012}$ & $\xms{1933.3781}{5.8987}$ & $\xms{3215.7598}{15.0527}$ & $\xms{1953.9355}{14.9148}$ & $\xms{0.1747}{0.0008}$ & $\xms{0.3569}{0.0017}$ & $\xms{0.1463}{0.0002}$ & $\xms{0.1960}{0.0026}$ & $\xms{0.7275}{0.0011}$ & $\xms{0.7923}{0.0021}$ & $\xms{0.6645}{0.0014}$ & $\xms{0.6221}{0.0050}$ & $\xms{0.7168}{0.0008}$ & $\xms{0.6425}{0.0005}$ \\ 
        ~ & RSF & $\xms{0.6407}{0.0044}$ & $\xms{0.6460}{0.0036}$ & $\xms{0.6099}{0.0039}$ & $\xms{1932.1591}{5.7560}$ & $\xms{3214.5663}{14.9170}$ & $\xms{1952.6491}{14.8421}$ & $\xms{0.1675}{0.0009}$ & $\xms{0.3442}{0.0014}$ & $\xms{0.1595}{0.0022}$ & $\xms{0.1782}{0.0031}$ & $\xms{0.6788}{0.0132}$ & $\xms{0.7510}{0.0058}$ & $\xms{0.6312}{0.0085}$ & $\xms{0.5947}{0.0026}$ & $\xms{0.7079}{0.0057}$ & $\xms{0.6374}{0.0037}$ \\ 
        ~ & GBSA & $\xms{0.6429}{0.0013}$ & $\xms{0.6429}{0.0013}$ & $\xms{0.6138}{0.0003}$ & $\xms{1934.7494}{5.8699}$ & $\xms{3216.8190}{15.0343}$ & $\xms{1954.4682}{14.9358}$ & $\xms{0.1827}{0.0005}$ & $\xms{0.3749}{0.0005}$ & $\xms{0.1287}{0.0010}$ & $\xms{0.2098}{0.0028}$ & $\xms{0.7115}{0.0049}$ & $\xms{0.8278}{0.0031}$ & $\xms{0.6581}{0.0020}$ & $\xms{0.6541}{0.0027}$ & $\xms{0.7049}{0.0041}$ & $\xms{0.6406}{0.0016}$ \\ 
        ~ & SurvivalTree & $\xms{0.5997}{0.0086}$ & $\xms{0.6034}{0.0076}$ & $\xms{0.5791}{0.0065}$ & - & - & - & $\xms{0.1689}{0.0023}$ & $\xms{0.3459}{0.0055}$ & $\xms{0.1637}{0.0050}$ & $\xms{0.1759}{0.0028}$ & $\xms{0.6188}{0.0131}$ & $\xms{0.7517}{0.0111}$ & $\xms{0.5938}{0.0105}$ & $\xms{0.5944}{0.0103}$ & $\xms{0.6469}{0.0103}$ & $\xms{0.5973}{0.0076}$ \\ 
        ~ & ExtraSurvivalTree & $\xms{0.5919}{0.0091}$ & $\xms{0.5967}{0.0081}$ & $\xms{0.5751}{0.0069}$ & $\xms{1947.5705}{0.0000}$ & $\xms{3259.7816}{0.0000}$ & $\xms{2000.2169}{0.0000}$ & $\xms{0.1711}{0.0031}$ & $\xms{0.3486}{0.0069}$ & $\xms{0.1544}{0.0074}$ & $\xms{0.1839}{0.0063}$ & $\xms{0.6073}{0.0144}$ & $\xms{0.7688}{0.0173}$ & $\xms{0.5911}{0.0095}$ & $\xms{0.6046}{0.0101}$ & $\xms{0.6345}{0.0099}$ & $\xms{0.5887}{0.0075}$ \\ 
        ~ & DeepHit & $\xms{0.4895}{0.0360}$ & $\xms{0.4941}{0.0031}$ & $\xms{0.4946}{0.0290}$ & $\xms{1937.1211}{5.9247}$ & $\xms{3219.1336}{15.0870}$ & $\xms{1956.6526}{14.9782}$ & $\xms{0.1985}{0.0007}$ & $\xms{0.4075}{0.0021}$ & $\xms{0.1030}{0.0003}$ & $\xms{0.2427}{0.0022}$ & $\xms{0.4957}{0.0244}$ & $\xms{0.9026}{0.0010}$ & $\xms{0.5082}{0.0180}$ & $\xms{0.7128}{0.0049}$ & $\xms{0.4792}{0.0416}$ & $\xms{0.4910}{0.0289}$ \\ 
        ~ & DeepSurv & $\xms{0.4922}{0.0465}$ & $\xms{0.4922}{0.0465}$ & $\xms{0.4975}{0.0319}$ & $\xms{57029.2069}{95344.1317}$ & $\xms{99678.2256}{166923.2737}$ & $\xms{139346.1680}{237745.4864}$ & $\xms{0.1660}{0.0221}$ & $\xms{0.3385}{0.0480}$ & $\xms{0.1673}{0.0483}$ & $\xms{0.1692}{0.0471}$ & $\xms{0.4840}{0.0871}$ & $\xms{0.7254}{0.1280}$ & $\xms{0.4910}{0.0583}$ & $\xms{0.5773}{0.1022}$ & $\xms{0.4848}{0.0666}$ & $\xms{0.4913}{0.0360}$ \\ 
        ~ & NMTLR & $\xms{0.5061}{0.0114}$ & $\xms{0.5063}{0.0109}$ & $\xms{0.5011}{0.0046}$ & $\xms{1934.7011}{5.2182}$ & $\xms{3218.8693}{13.7462}$ & $\xms{1957.9926}{14.6666}$ & $\xms{0.1832}{0.0158}$ & $\xms{0.3757}{0.0339}$ & $\xms{0.1303}{0.0285}$ & $\xms{0.2090}{0.0340}$ & $\xms{0.5219}{0.0316}$ & $\xms{0.8253}{0.0817}$ & $\xms{0.5146}{0.0259}$ & $\xms{0.5153}{0.0161}$ & $\xms{0.6518}{0.0621}$ & $\xms{0.5057}{0.0107}$ \\ 
        ~ & DeepCoxPH & $\xms{0.6460}{0.0008}$ & $\xms{0.6460}{0.0008}$ & $\xms{0.6147}{0.0009}$ & $\xms{919.4983}{6.5921}$ & $\xms{2077.0564}{8.7011}$ & $\xms{1954.4660}{12.0548}$ & $\xms{0.0689}{0.0003}$ & $\xms{0.1614}{0.0017}$ & $\xms{0.6611}{0.0004}$ & $\xubms{0.0001}{0.0001}$ & $\xms{0.7262}{0.0006}$ & $\xms{0.0699}{0.0012}$ & $\xms{0.6647}{0.0017}$ & $\xms{0.1856}{0.0005}$ & $\xms{0.7155}{0.0008}$ & $\xms{0.6425}{0.0003}$ \\ 
        ~ & DSM-Weibull & $\xms{0.4713}{0.0000}$ & $\xms{0.4807}{0.0000}$ & $\xms{0.4876}{0.0000}$ & $\xms{1931.6013}{0.0000}$ & $\xms{3204.5515}{0.0000}$ & $\xms{1941.2053}{0.0000}$ & $\xms{0.2014}{0.0000}$ & $\xms{0.4144}{0.0000}$ & $\xms{0.0990}{0.0000}$ & $\xms{0.2480}{0.0000}$ & $\xms{0.4292}{0.0000}$ & $\xms{0.9181}{0.0000}$ & $\xms{0.4844}{0.0000}$ & $\xms{0.7213}{0.0000}$ & $\xms{0.4267}{0.0000}$ & $\xms{0.4750}{0.0000}$ \\ 
        ~ & DSM-LogNormal & $\xms{0.4596}{0.0024}$ & $\xms{0.4460}{0.0010}$ & $\xms{0.4956}{0.0044}$ & $\xms{1936.8245}{5.9554}$ & $\xms{3218.9133}{15.1108}$ & $\xms{1956.4963}{15.0003}$ & $\xms{0.2020}{0.0008}$ & $\xms{0.4148}{0.0023}$ & $\xms{0.0996}{0.0007}$ & $\xms{0.2497}{0.0023}$ & $\xms{0.3408}{0.0008}$ & $\xms{0.9179}{0.0015}$ & $\xms{0.4262}{0.0037}$ & $\xms{0.7258}{0.0053}$ & $\xms{0.3611}{0.0004}$ & $\xms{0.4659}{0.0016}$ \\ 
        ~ & DeepCoxMixtures & $\xbms{0.6488}{0.0006}$ & $\xbms{0.6522}{0.0003}$ & $\xbms{0.6164}{0.0007}$ & $\xbms{909.8300}{7.6157}$ & $\xbms{2061.4934}{9.1529}$ & $\xbms{1924.2608}{16.0933}$ & $\xbms{0.0679}{0.0004}$ & $\xbms{0.1607}{0.0015}$ & $\xbms{0.6636}{0.0004}$ & $\xbms{0.0005}{0.0000}$ & $\xbms{0.7417}{0.0020}$ & $\xbms{0.0671}{0.0010}$ & $\xbms{0.6665}{0.0014}$ & $\xbms{0.1849}{0.0005}$ & $\xbms{0.7226}{0.0006}$ & $\xbms{0.6446}{0.0005}$ \\ 
        ~ & SurvivalPFN & $\xms{0.6315}{0.0013}$ & $\xms{0.6321}{0.0010}$ & $\xms{0.5984}{0.0025}$ & $\xms{932.0172}{7.8315}$ & $\xms{2147.2668}{7.5478}$ & $\xms{1950.5671}{14.6757}$ & $\xms{0.0692}{0.0003}$ & $\xms{0.1661}{0.0015}$ & $\xms{0.6594}{0.0004}$ & $\xms{0.0011}{0.0002}$ & $\xms{0.7168}{0.0026}$ & $\xms{0.0714}{0.0012}$ & $\xms{0.6535}{0.0007}$ & $\xms{0.1915}{0.0006}$ & $\xms{0.7010}{0.0019}$ & $\xms{0.6288}{0.0002}$ \\ 
        ~ & \textbf{EnSurv} & $\xubms{0.6625}{0.0008}$ & $\xubms{0.6594}{0.0019}$ & $\xubms{0.6229}{0.0014}$ & $\xubms{854.7873}{12.7169}$ & $\xubms{2002.1295}{1.9120}$ & $\xubms{1706.2486}{72.1885}$ & $\xubms{0.0670}{0.0012}$ & $\xubms{0.1596}{0.0018}$ & $\xubms{0.6638}{0.0061}$ & $\xms{0.0007}{0.0000}$ & $\xubms{0.7543}{0.0042}$ & $\xubms{0.0642}{0.0009}$ & $\xubms{0.6782}{0.0006}$ & $\xubms{0.1839}{0.0011}$ & $\xubms{0.7459}{0.0039}$ & $\xubms{0.6599}{0.0005}$ \\ \bottomrule

\end{tabular}
\end{adjustbox}
\end{table}

\subsection{Model Explainability}
EnSurv possesses built-in explainability analysis tools via descriptive parameter summaries. We derive what the generative neural network layers learn about the clinical and region-level covariates from the dataset, that influence graft survival times in patients via evaluating both the linear and non-linear network capacities allocated to each covariate. The raw linear effect, denoted as $\beta_i$ for the $i^{th}$ feature, is extracted directly from the learned parameters of the linear layer. 
The non-linear contribution $\gamma_i$ 
is calculated by tracing the structural path weights across the network, defined as the $L_2$ norm of the $i^{th}$ column of the accumulated path matrix $P$, where $P = \prod_{l=1}^L \vert{}W^{(l)}\vert{}$ represents the product of the absolute weight matrices of all sequential fully connected layers. Finally, the total importance for each feature is aggregated as the sum of its absolute linear magnitude and non-linear contribution, 
which is subsequently normalized to yield the relative total importance percentage. 

\begin{figure}[t]
    \centering
    \includegraphics[width=\textwidth]{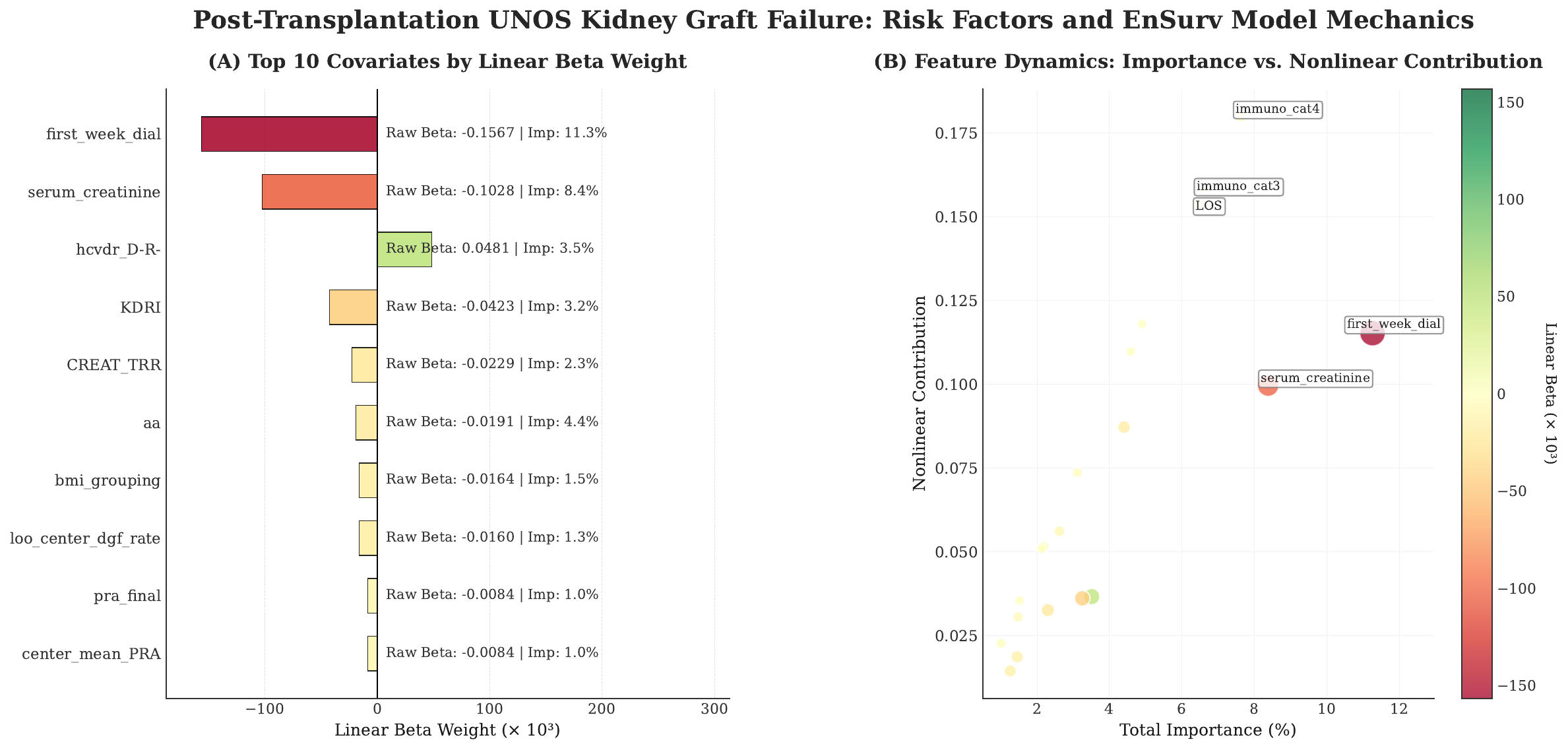}
    \caption{Explainability analysis of the UNOS data (average of 3 folds) via the trained EnSurv model in the post-transplantation setting.}
    \label{fig:feature_importance}
\end{figure}

Overall, EnSurv assigns $\sim 84.1\%$ importance to patient-level variables and the rest to region-level covariates. Figure~\ref{fig:feature_importance}(A) highlights the ten patient-level covariates identified by the model as exerting the greatest influence on graft failure with respect to the coefficients ($\beta$). A negative weight for a covariate indicates that negative association was learnt for that variable. For example, EnSurv identifies the necessity for dialysis within the first week of transplant, increased creatinine levels during discharge, an elevated KDRI, high recipient BMI, the requirement for pre-transplant dialysis, pre-existing peripheral vascular disease (PVD), and high panel reactive antibody (PRA) scores as critical indicators of post-transplant premature graft loss. 

These model-derived associations align with established clinical mechanisms. For instance, pre-existing PVD in transplant candidates introduces severe anatomical and hemodynamic complications 
that directly exacerbate the risk of acute and early graft failure \citep{hernandez2021peripheral}. Likewise, an elevated PRA score indicates a highly sensitized recipient immune system, widely recognized to predispose patients to acute rejection episodes and subsequent graft loss \citep{oweira2022risk}. Finally, the model's association between African American recipient race and an increased hazard of early graft failure corroborates the demographic disparities previously documented in broad transplantation studies \citep{taber2014significant}. EnSurv further learns a positive effect of the engineered immuno-entropy on graft survival. When holding baseline patient risk constant, institutions in regions that adapt their pharmacological protocols to the patient (higher entropy) are observed to outperform institutions in terms of longer graft survival that rigidly apply standardized regimens. Figure~\ref{fig:feature_importance}(B) presents the total feature importance versus the contribution of the nonlinear layers, illustrating how both the linear and nonlinear components of EnSurv influences the total covariate importance.

\section{Discussion}\label{sec:conclusion}
We proposed EnSurv, a deep generative framework for predicting time to kidney graft failure from right-censored transplant records. Our theoretical results identify the conditional distribution targeted by pooled completion, bound its discrepancy from the true restricted graft-survival distribution, and establish consistency for the best attainable risk within the pre-ANM family under the stated sampling and optimization conditions. Across three UNOS validation folds, EnSurv achieved the highest C-indices among the methods evaluated, both before transplantation ($0.609$) and after transplantation ($0.663$), along with lowest Brier scores. The model's linear pathway provides a descriptive summary of fitted associations, while its gradient penalty allows prespecified clinical knowledge to inform the fit.

\subsection*{Limitations and Future Directions}
First, the use of a pooled KM tail is a deliberate compromise. Pooling allows the completion of censored records to draw on the full cohort still at risk, which is particularly valuable late in follow-up when few recipients with similar clinical profiles remain under observation. Recipients censored at the same time therefore share a completion distribution, even when their subsequent graft prognosis differs. A natural extension is to compare pooled completion with tails estimated within clinically defined strata or by weighting recipients with similar covariates. Under conditional independent censoring and adequate follow-up within covariate profiles, consistent estimation of these conditional tails would remove the discrepancy attributable to pooling. Calibration within donor-risk strata would provide a practical way to assess whether the additional tailoring improves prediction.

Several directions for future development remain. The current study focuses on graft loss as the primary event; however, patient death with a functioning graft operates as a critical competing risk \citep{coemans2025competing}. Extending the EnSurv framework to handle competing risks is a primary future direction. Second, future work could incorporate region-specific `random effects' to more accurately model the hierarchical, multicenter structure of national transplant registries. Finally, explicitly modeling the correlation structures within specific clinical subpopulations, such as cohorts defined by Donor/Recipient HCV status, could further refine the precision of the survival estimates for highly specialized patient profiles.

\section{Conflicts of Interest}
The authors declare that they have no competing interests.


\section{Supplementary Material}
The supplementary material consists of additional theoretical results, mathematical proofs, description of the evaluation metrics, and quantitative results.

\section{Data \& Code Availability}
All code and simulation datasets used in this study are publicly available at \url{https://github.com/PyCoder913/EnSurv}. The UNOS kidney transplantation dataset is not publicly available and may be obtained from the corresponding author upon reasonable request.


\section{Acknowledgments}
The authors thank Mr. Rahul Goswami, PhD scholar at Indian Institute of Technology Guwahati, for participating in fruitful discussions.


\clearpage
\bibliographystyle{oup-abbrvnat}
\bibliography{reference}




\renewcommand{\thetable}{S.\arabic{table}}
\renewcommand{\thefigure}{S.\arabic{figure}}
\setcounter{table}{0}
\setcounter{figure}{0}

\setcounter{theorem}{0}
\renewcommand{\thetheorem}{S.\arabic{theorem}}

\setcounter{equation}{0}
\renewcommand{\theequation}{S.\arabic{equation}}
\setcounter{proposition}{0}
\renewcommand{\theproposition}{S.\arabic{proposition}}

\clearpage

\section*{Supplementary Material}

\begin{appendices}

\section{Additional Theoretical Results}\label{appendix:theory}
In this section, we discuss additional theoretical properties of our proposed model for survival analysis. We begin by expressing the effect of pooled completion on the survival probability at a given time $t<\tau$. Proposition~\ref{prop:bias-identity} gives an exact expression for this difference in terms of censoring before $t$. Lemmas~\ref{lem:rep} and~\ref{lem:km} provide the two ingredients behind the completion-loss rate in Theorem~\ref{thm:psi}: the first controls how the loss changes with the survival curve, and the second uses uniform empirical distribution bounds \citep{massart1990tight} to control estimation of the pooled curve and its bootstrap replicates. We then turn to the fitted survival probabilities. Theorem~\ref{thm:dist} identifies their limit at a given covariate profile, and Corollary~\ref{cor:T} gives sufficient conditions for that limit to equal the true graft-survival probability at continuity points below $\tau$. Finally, Proposition~\ref{prop:excess} bounds prediction error in terms of model approximation, estimation, optimization, and pooling, and quantifies how much pooled completion can change the expected duration of graft function up to $\tau$. The proofs are given in Appendix~\ref{appendix:proofs}.

\begin{proposition}[Survival distortion]\label{prop:bias-identity}
Under conditional independent censoring, for $0<t<\tau$,
\begin{equation}\label{eq:bias-identity}
1-F_x^*(\log t)-S_T(t\mid x)
=\int_{[c_0,t)}\left\{S_T(c\mid x)\frac{S^\#(t)}{S^\#(c)}-S_T(t\mid x)\right\}P_{C\mid x}(dc).
\end{equation}
The integral is zero for $t\le c_0$.
\end{proposition}

This proposition establishes an exact, integral representation of the prognostic bias introduced when patient-specific unobserved lifetimes are imputed using a population-level pooled survival tail.

\begin{lemma}[Residual representation and stability]\label{lem:rep}
Let $W$ take values in $(0,\tau]$ and have survival function $S(t)=\mathbb P(W>t)$. For $c\in[c_0,\tau)$ with $S(c)>0$ and $u\in\mathbb R$, define $\Psi_S(c,u)=\mathbb E[|\log W-u|\mid W>c]$. Then
\begin{equation}\label{eq:rep}
\Psi_S(c,u)=|\log c-u|+\int_c^\tau\operatorname{sgn}(\log t-u)\frac{S(t)}{S(c)}\frac{dt}{t}.
\end{equation}
It is 1-Lipschitz in $u$ and bounded above by $|u|+\max\{|\log c_0|,|a|\}$. If $S_0(c)\ge s>0$ on $[c_0,\tau)$ and $e:=\|S-S_0\|_{\infty,[c_0,\tau)}\le s/2$, then
\begin{equation}\label{eq:tail-stability}
\sup_{c_0\le c<\tau,\,u\in\mathbb R}|\Psi_S(c,u)-\Psi_{S_0}(c,u)|\le\frac{4L}{s}e.
\end{equation}
Here $\|f\|_{\infty,I}=\sup_{t\in I}|f(t)|$ and $\operatorname{sgn}$ is the sign function, with value zero at zero.
\end{lemma}

Lemma~\ref{lem:rep} isolates the expected residual loss into a closed-form integral that avoids explicitly tracking individual jump locations and masses, enabling stable and continuous bounds on how estimation errors in the pooled KM curve uniformly propagate into the final objective function.

\begin{lemma}[Pooled KM and bootstrap convergence]\label{lem:km}
Let $\widehat S_n$ be the ordinary pooled KM estimate and $\widehat S_n^{(b)}$ a size-$n$ bootstrap estimate. With the supremum over $0\le t<\tau$,
\begin{equation}\label{eq:km-rate}
\mathbb E\left\|\widehat S_n-S^\#\right\|_\infty=O\left(n^{-1/2}\right),\qquad
\mathbb E\left\|\widehat S_n^{(b)}-S^\#\right\|_\infty=O\left(n^{-1/2}\right).
\end{equation}
Moreover, $S^\#(\tau-)\ge h_0$. The expectations include sampling and for the second bound, bootstrap randomness.
\end{lemma}

Hence, the empirical KM estimator and its bootstrap replicates estimate the pooled working tail at a parametric $n^{-1/2}$ rate, uniformly across the entire observation window.

\begin{theorem}[Survival convergence at a covariate profile]\label{thm:dist}
Under Theorem~\ref{thm:argmin}, fix $x$ such that $\nu\mapsto U_\nu(x,\epsilon)$ is continuous for $P_\epsilon$-almost every $\epsilon$. Suppose every $\nu_0\in\Theta_0$ induces the same law $Q_{0,x}$ at this profile. Then, for $0<t<\tau$ with $Q_{0,x}(\{\log t\})=0$,
\begin{equation}\label{eq:survival-convergence}
1-F_{\widehat\nu_n,x}(\log t)\xrightarrow{P}Q_{0,x}((\log t,\infty)).
\end{equation}
\end{theorem}

Theorem~\ref{thm:dist} guarantees that at profiles satisfying the continuity and common-law conditions, fitted survival probabilities converge at continuity points to those of the common risk-minimizing predictive law.

\begin{corollary}\label{cor:T}
Under Theorem~\ref{thm:argmin}, suppose some $\nu^*\in\Theta$ realizes $P_x^*$ for almost every $x$. Then all minimizers induce $P_x^*$ on a common full-measure covariate set, and Theorem~\ref{thm:dist} applies there with $Q_{0,x}=P_x^*$. If~\eqref{eq:positivity} and the residual agreement in Proposition~\ref{prop:pooling} also hold, the limit in~\eqref{eq:survival-convergence} is $S_T(t\mid x)$ at its continuity points below $\tau$. Using $M=M_n\to\infty$ independent prediction draws in~\eqref{eq:prediction-mc} preserves these conclusions.
\end{corollary}


\begin{proposition}[Prediction error and restricted mean survival]\label{prop:excess}
Under Theorem~\ref{thm:argmin}, define the model approximation error
\[
a_\Theta=\inf_{\nu\in\Theta}\mathbb E_X\int_{\mathbb R}\{F_{\nu,X}(z)-F_X^*(z)\}^2\,dz.
\]
Then
\begin{equation}\label{eq:excess-pseudo}
\mathbb E_X\int_{\mathbb R}\{F_{\widehat\nu_n,X}(z)-F_X^*(z)\}^2\,dz
\le a_\Theta+2\Delta_n+r_n.
\end{equation}
If $T\perp C\mid X$, let $\bar b=\mathbb E_X b(X)\le L\mathbb P(D=0)$. Then
\begin{equation}\label{eq:true-cdf-bound}
\mathbb E_X\int_{\mathbb R}\{F_{\widehat\nu_n,X}(z)-F_X(z)\}^2\,dz
\le2a_\Theta+4\Delta_n+2r_n+2\bar b.
\end{equation}
Here $\mathbb E_X$ integrates a fresh covariate draw, conditional on the fitted model. For the population targets at a fixed profile,
\begin{equation}\label{eq:time-wasserstein}
\begin{split}
W_1\{\mathcal L(e^{Z^*}\mid X=x),\mathcal L(T\wedge\tau\mid X=x)\}&\le\tau b(x),\\
|\mathbb E(e^{Z^*}\mid X=x)-\mathbb E(T\wedge\tau\mid X=x)|&\le\tau b(x).
\end{split}
\end{equation}
\end{proposition}

These bounds decouple the overall error of EnSurv into components of model approximation, sampling and simulation variation, optimization inaccuracy, and pooling discrepancy, with only the tail-estimation component achieving the previously established $n^{-1/2}$ convergence rate. Furthermore, the original-time expectation in~\eqref{eq:time-wasserstein} precisely defines the restricted mean survival time, $\mathbb E(T\wedge\tau\mid X=x)=\int_0^\tau S_T(t\mid x)\,dt$. While this yields a clinically interpretable expected duration of graft function within the observation window, it does not extrapolate to unrestricted life expectancy. Collectively, this theoretical framework ensures a transparent characterization of both the working predictor and its inherent structural limitations.

\section{Mathematical Proofs}\label{appendix:proofs}
\subsection{Proof for Proposition~\ref{prop:target}}
\begin{proof}
The construction of $Z^*$ gives
\[
\mathbb E(|Z^*-u|\mid Y,\delta,X)=D|A-u|+(1-D)\Psi(Y,u).
\]
Take the conditional expectation given $X=x$ and integrate over $u\sim Q$. Tonelli's theorem and the finite first moments justify these operations and the subtraction of the dispersion term, proving~\eqref{eq:target-risk}. The identity
\[
|v-u|=\int_{\mathbb R}|\mathbf1\{v\le z\}-\mathbf1\{u\le z\}|\,dz
\]
applied to the three independent pairwise distances in $r_x(Q)-r_x(P_x^*)$ yields~\eqref{eq:energy-identity}. If the integral vanishes, the CDFs agree Lebesgue-almost everywhere and hence everywhere by right continuity. Finally, realizability supplies a model attaining the integral of $r_x(P_x^*)$. Every minimizer must therefore have zero integrated nonnegative CDF discrepancy, which implies equality of laws for $P_X$-almost every $x$.
\end{proof}

\subsection{Proof for Proposition~\ref{prop:pooling}}
\begin{proof}
Throughout the proof, fix a profile $x$ for which the stated
conditional assumptions hold.

\medskip
\noindent (i) First, note that
\(
\{D=0\}=\{T>C,\ C<\tau\}.
\)
Consequently, conditional independence gives
\[
\mathbb P(C\in dc,\ D=0\mid X=x)
=
\mathbf 1\{c<\tau\}
S_T(c\mid x)\,P_{C\mid x}(dc).
\]
In particular, for censoring times with positive weight,
\[
\mathcal L(Z_\tau\mid C=c,D=0,X=x)=R_x(c).
\]
The pseudo-completion instead uses $K_c$. On $D=1$, both outcomes
equal the observed restricted log-time. Therefore, for every Borel
set $E\subseteq\mathbb R$,
\begin{align*}
P_x(E)
&=
\mathbb P(Z_\tau\in E,D=1\mid X=x)
+
\int_{[c_0,\tau)}
S_T(c\mid x)R_x(c)(E)\,P_{C\mid x}(dc),\\
P_x^*(E)
&=
\mathbb P(Z_\tau\in E,D=1\mid X=x)
+
\int_{[c_0,\tau)}
S_T(c\mid x)K_c(E)\,P_{C\mid x}(dc).
\end{align*}
These identities exhibit the common observed component and the
two different residual components. Both residual laws are supported on $(\log c,a]$. Hence, the
pseudo-complete law also has a finite first moment:
\[
\mathbb E(|Z^*|\mid X=x)
\le
\mathbb E(|Z_\tau|\mid X=x)
+\max\{|\log c_0|,|a|\}
<\infty.
\]
Thus, the Wasserstein distances below are well-defined.

For a probability law $P$ on $\mathbb R$, let
\[
F_P^{-1}(r)=\inf\{z:F_P(z)\ge r\},
\qquad 0<r<1,
\]
be its generalized inverse CDF. We briefly verify the optimality of
the common-uniform coupling. For any coupling $(V,U)$ of $P,Q$,
Tonelli's theorem and the triangle inequality give
\begin{align*}
\mathbb E|V-U|
&=
\int_{\mathbb R}
\mathbb E\left|
\mathbf 1\{V\le z\}-\mathbf 1\{U\le z\}
\right|\,dz\\
&\ge
\int_{\mathbb R}|F_P(z)-F_Q(z)|\,dz.
\end{align*}
If $\xi\sim\operatorname{Unif}(0,1)$, the quantile coupling attains
this lower bound, since
\begin{align*}
\mathbb E\left|F_P^{-1}(\xi)-F_Q^{-1}(\xi)\right|
&=
\int_{\mathbb R}\int_0^1
\left|
\mathbf 1\{r\le F_P(z)\}
-
\mathbf 1\{r\le F_Q(z)\}
\right|\,dr\,dz\\
&=
\int_{\mathbb R}|F_P(z)-F_Q(z)|\,dz\\
&=W_1(P,Q).
\end{align*}

Now take a uniform variable $\xi$, independent of the observed
data, and construct a pair $(V,U)$ by
\[
(V,U)=
\begin{cases}
(Z_\tau,Z_\tau),&D=1,\\[1ex]
\bigl(F_{R_x(C)}^{-1}(\xi),F_{K_C}^{-1}(\xi)\bigr),&D=0.
\end{cases}
\]
The residual laws are measurable probability kernels in $c$, and
their generalized inverses have measurable versions. Thus this
construction defines a valid coupling. The mixture identities above
show that, conditional on $X=x$, its marginals are $P_x$ and $P_x^*$.

The coupling has zero cost on $D=1$, and its conditional cost at
$C=c,D=0$ is $W_1\{R_x(c),K_c\}$. Therefore,
\begin{align*}
W_1(P_x^*,P_x)
&\le \mathbb E(|U-V|\mid X=x)\\
&=
\int_{[c_0,\tau)}
S_T(c\mid x)W_1\{R_x(c),K_c\}\,P_{C\mid x}(dc)\\
&=b(x).
\end{align*}
Because both residual laws are supported on $(\log c,a]$, any pair
of their possible values differs by at most $a-\log c$. In particular,
\[
W_1\{R_x(c),K_c\}
\le a-\log c
=\log(\tau/c)
\le L.
\]
It follows that
\begin{align*}
b(x)
&\le
L\int_{[c_0,\tau)}
S_T(c\mid x)\,P_{C\mid x}(dc)\\
&=
L\,\mathbb P(T>C,\ C<\tau\mid X=x),
\end{align*}
which proves both inequalities in~\eqref{eq:pooling-bound}.

\medskip
\noindent(ii)
Subtracting the two mixture identities yields, for every Borel set
$E$,
\[
P_x^*(E)-P_x(E)
=
\int_{[c_0,\tau)}
S_T(c\mid x)
\{K_c(E)-R_x(c)(E)\}\,P_{C\mid x}(dc).
\]
Under residual agreement, the integrand vanishes at almost every
censoring time with positive weight. Where $S_T(c\mid x)=0$, its
contribution is also zero. Hence
\[
P_x^*(E)=P_x(E)
\qquad\text{for every Borel set }E,
\]
and therefore $P_x^*=P_x$.

\medskip
\noindent(iii)
Stochastic dominance of $K_c$ over $R_x(c)$ means
\(
F_{K_c}(z)\le F_{R_x(c)}(z)\)
for every $z\in\mathbb R$.
Applying the mixture identities to $E=(-\infty,z]$ gives
\[
F_x^*(z)-F_x(z)
=
\int_{[c_0,\tau)}
S_T(c\mid x)
\{F_{K_c}(z)-F_{R_x(c)}(z)\}\,P_{C\mid x}(dc)
\le0.
\]
Thus, $P_x^*$ stochastically dominates $P_x$. The same calculation
with the opposite inequality proves the reversed order.

It remains to verify the sufficient hazard condition. For
$c_0\le c<t<\tau$, the residual survival probabilities are
\[
1-F_{R_x(c)}(\log t)
=
\frac{S_T(t\mid x)}{S_T(c\mid x)},
\qquad
1-F_{K_c}(\log t)
=
\frac{S^\#(t)}{S^\#(c)}.
\]
The corresponding product-integral representations are
\[
\frac{S_T(t\mid x)}{S_T(c\mid x)}
=
\prod_{(c,t]}(1-d\Lambda_x),
\qquad
\frac{S^\#(t)}{S^\#(c)}
=
\prod_{(c,t]}(1-d\Lambda^\#).
\]

To make their comparison explicit, denote the non-atomic part of
each hazard measure by $d\Lambda^{\mathrm{cont}}$ and its jump at
$s$ by
$\Delta\Lambda(s)=\Lambda(s)-\Lambda(s-)$.
The non-atomic part may include a singular continuous component.
For either hazard measure,
\[
\prod_{(c,t]}(1-d\Lambda)
=
\exp\left\{-\int_{(c,t]}d\Lambda^{\mathrm{cont}}\right\}
\prod_{c<s\le t}\{1-\Delta\Lambda(s)\}.
\]
Measure domination implies
\[
d\Lambda_x^{\mathrm{cont}}
\ge d(\Lambda^\#)^{\mathrm{cont}},
\qquad
\Delta\Lambda_x(s)\ge\Delta\Lambda^\#(s)
\quad\text{for every }s\in(c_0,\tau).
\]
The product can be taken over the union of the two sets of jump
times, with a zero jump contributing a factor of one.
Positivity of the survivals before $\tau$ ensures that these
factors are positive. Comparing the exponential terms and jump
factors therefore gives
\[
\frac{S_T(t\mid x)}{S_T(c\mid x)}
\le
\frac{S^\#(t)}{S^\#(c)}
\qquad(c<t<\tau).
\]
Equivalently,
\[
F_{K_c}(\log t)\le F_{R_x(c)}(\log t)
\qquad(c<t<\tau).
\]

For $z\le\log c$, both residual CDFs are zero, and for $z\ge a$,
both are one. Thus the CDF inequality holds on the entire real line.
The terminal atoms are also included: taking $t\uparrow\tau$ gives
\[
R_x(c)(\{a\})
=
\frac{S_T(\tau-\mid x)}{S_T(c\mid x)}
\le
\frac{S^\#(\tau-)}{S^\#(c)}
=
K_c(\{a\}).
\]
Here the true terminal mass includes failures exactly at $\tau$
as well as later failures, because $Z_\tau=a$ whenever $T\ge\tau$.

If the hazard measures are equal on $(c_0,\tau)$, all the displayed
residual survival ratios are equal. Together with the common
support boundaries, this implies $R_x(c)=K_c$, proving the
sufficient condition for part~(ii). Reversing the hazard-measure
inequality reverses every survival comparison and hence the
stochastic order.
\end{proof}

\subsection{Proof for Proposition~\ref{prop:bias-identity}}
\begin{proof}
On $C\ge t$, $Z^*>\log t$ if and only if $T>t$, including when $C=t$. Conditional independence makes this contribution $G(t\mid x)S_T(t\mid x)$. When $C=c<t$, only $T>c$ can contribute; its completion exceeds $\log t$ with probability $S^\#(t)/S^\#(c)$. Thus
\[
1-F_x^*(\log t)=G(t\mid x)S_T(t\mid x)
+\int_{[c_0,t)}S_T(c\mid x)\frac{S^\#(t)}{S^\#(c)}P_{C\mid x}(dc).
\]
Subtract $S_T(t\mid x)$ and use $1-G(t\mid x)=\mathbb P(C<t\mid X=x)$.
\end{proof}

\subsection{Proof for Lemma~\ref{lem:rep}}
\begin{proof}
For a time $W$ drawn conditionally above $c$, absolute continuity of $t\mapsto|\log t-u|$ gives
\[
|\log W-u|=|\log c-u|+\int_c^\tau\operatorname{sgn}(\log t-u)\mathbf1\{W>t\}\frac{dt}{t}.
\]
Fubini applies since the integral is bounded by $L$. Taking expectations gives~\eqref{eq:rep}; the Lipschitz and envelope bounds follow from the absolute-value inequality and the bounded residual support. Since $S(c)\ge s/2$ and $S_0(t)\le S_0(c)$ for $t\ge c$,
\[
\left|\frac{S(t)}{S(c)}-\frac{S_0(t)}{S_0(c)}\right|
\le\frac{|S(t)-S_0(t)|}{S(c)}+
\frac{S_0(t)|S_0(c)-S(c)|}{S(c)S_0(c)}\le\frac{4e}{s}.
\]
Subtract~\eqref{eq:rep} for the two laws and integrate $dt/t$ to obtain~\eqref{eq:tail-stability}.
\end{proof}

\subsection{Proof for Lemma~\ref{lem:km}}
\begin{proof}
Let $H_n(t)=n^{-1}\sum_i\mathbf1\{Y_i\ge t\}$ and $N_n(t)=n^{-1}\sum_i\mathbf1\{Y_i\le t,\delta_i=1\}$. Empirical-CDF bounds give
\[
\mathbb E e_n=O(n^{-1/2}),\qquad
e_n=\|H_n-H\|_\infty+\|N_n-N\|_\infty.
\]
These bounds also cover the subdistribution $N_n$ by assigning nonfailures a value beyond the window, and cover $H_n$ through CDF left limits. On $\|H_n-H\|_\infty\le h_0/2$, define $\Lambda_n(t)=\int_{(0,t]}H_n^{-1}\,dN_n$. The decomposition
\[
\Lambda_n-\Lambda^\#=\int(H_n^{-1}-H^{-1})\,dN_n+
\int H^{-1}\,d(N_n-N)
\]
and Stieltjes integration by parts give
\[
\|\Lambda_n-\Lambda^\#\|_\infty
\le 2h_0^{-2}\|H_n-H\|_\infty+2h_0^{-1}\|N_n-N\|_\infty,
\qquad \Lambda_n(\tau-)\le2/h_0.
\]
The integration-by-parts bound uses that $H^{-1}$ is bounded, nondecreasing, and left-continuous; the right-continuous integrator retains its full jumps, so ties are allowed.

Subtract the two survival equations $S(t)=1-\int_{(0,t]}S(s-)\,d\Lambda(s)$. Integration by parts bounds the forcing term $\int S^\#(s-)\,d(\Lambda_n-\Lambda^\#)(s)$ by $2\|\Lambda_n-\Lambda^\#\|_\infty$. Iterating the remaining Volterra inequality over strictly ordered integration times yields
\[
\|\widehat S_n-S^\#\|_\infty
\le2\exp\{\Lambda_n(\tau-)\}\|\Lambda_n-\Lambda^\#\|_\infty.
\]
(The iterated integrals of order $k$ are at most $\Lambda_n(\tau-)^k/k!$.) Off this event, the survival error is at most one; Markov's inequality gives an exceptional-event probability at most $2\mathbb E e_n/h_0$. This proves the first expected rate.

For a bootstrap sample, conditional empirical-CDF bounds around $H_n,N_n$ give expected supremum error $O(n^{-1/2})$ with a distribution-free constant. The triangle inequality therefore gives
\[
\mathbb E\{\|H_n^{(b)}-H\|_\infty+\|N_n^{(b)}-N\|_\infty\}=O(n^{-1/2}).
\]
Apply the same deterministic hazard and survival bounds on $\|H_n^{(b)}-H\|_\infty\le h_0/2$, and the same exceptional-event argument, to prove the bootstrap rate.

Finally, at an observed time with $r$ at risk, $d$ failures, and $k$ censorings, the KM multiplier $1-d/r$ is at least the all-removals multiplier $1-(d+k)/r$. Multiplying before $\tau$ gives $\widehat S_n(\tau-)\ge H_n(\tau)$. Uniform convergence and the ordinary law of large numbers imply $S^\#(\tau-)\ge H(\tau)=h_0$.
\end{proof}

\subsection{Proof for Theorem~\ref{thm:psi}}
\begin{proof}
Put $s_0=S^\#(\tau-)\ge h_0$ and $e_{n,b}=\|\widehat S_n^{(b)}-S^\#\|_\infty$. When $e_{n,b}\le s_0/2$, Lemma~\ref{lem:rep} bounds replicate $b$'s uniform residual error by $4Le_{n,b}/s_0$. Otherwise, both tails, including the fallback, are probability laws on $[\log c,a]$. The difference of their expected $|z-u|$ is at most $L$, uniformly in $u$. The expected residual error is therefore at most
\[
\frac{4L}{s_0}\mathbb E e_{n,b}+L\mathbb P(e_{n,b}>s_0/2)
\le\frac{6L}{s_0}\mathbb E e_{n,b}=O(n^{-1/2}).
\]
The error of an average is at most the average of the errors, so the same expectation bound holds for every $B_n$. Markov's inequality gives the probability rate. Replicate independence is unnecessary for this argument, and the proof is complete.
\end{proof}

\subsection{Proof for Theorem~\ref{thm:ulln}}
\begin{proof}
Let $\widetilde R_n$ be~\eqref{eq:emp-loss} with $\Psi$ in place of $\widehat\Psi_n$. Pathwise,
\begin{equation}\label{eq:plugin-bound}
\sup_{\nu\in\Theta}\left|\widehat R_n(\nu)-\widetilde R_n(\nu)\right|\le\eta_n\xrightarrow{P}0.
\end{equation}
No sample splitting is needed for this bound. Let $W_i=(Y_i,\delta_i,X_i,\epsilon_{i1},\ldots,\epsilon_{im})$ and denote the ideal criterion's summand by $f_\nu(W_i)$. These extended observations are i.i.d., and the distinct-pair normalization gives $\mathbb E f_\nu(W_i)=R(\nu)$. Parameter continuity and the Lipschitz property of $\Psi$ make $f_\nu$ almost surely continuous in $\nu$. Its integrable envelope is
\begin{equation}\label{eq:loss-envelope}
\sup_{\nu\in\Theta}|f_\nu(W_i)|\le D_i|A_i|+(1-D_i)\max\{|\log c_0|,|a|\}
+\frac2m\sum_{j=1}^m\sup_{\nu\in\Theta}|U_\nu(X_i,\epsilon_{ij})|.
\end{equation}
Here $D_i|A_i|=D_i|Z_{\tau,i}|$, so~\eqref{eq:cond_1} and (A3) ensure integrability. Dominated convergence proves finiteness and continuity of $R$.

For each fixed $\nu$, continuity and domination give
\[
\lim_{r\downarrow0}\mathbb E\left[\sup_{\mu\in\Theta:\|\mu-\nu\|<r}|f_\mu-f_\nu|\right]=0.
\] Compactness supplies a finite cover by neighborhoods with arbitrarily small expected oscillations. The strong law at their centers and for their oscillation envelopes then gives uniform almost-sure convergence of $\widetilde R_n$ to $R$. These suprema are measurable by separability and almost-sure continuity; a countable sequence of covers suffices. Combine this convergence with~\eqref{eq:plugin-bound} to obtain~\eqref{eq:uniform-loss}.
\end{proof}

\subsection{Proof for Theorem~\ref{thm:argmin}}
\begin{proof}
Under the assumptions of Theorem~\ref{thm:ulln}, the parameter
space $\Theta$ is nonempty and compact, the population risk
$R:\Theta\to\mathbb R$ is continuous, and
\(
\Delta_n
:=
\sup_{\nu\in\Theta}
\left|\widehat R_n(\nu)-R(\nu)\right|
\xrightarrow{P}0.
\)
We assume explicitly that the measurable estimator
$\widehat\nu_n$ takes values in $\Theta$.

By continuity of $R$ on the nonempty compact set $\Theta$,
the minimum
$R_{\min}:=\min_{\nu\in\Theta}R(\nu)$ is attained. Consequently, $\Theta_0=\{\nu\in\Theta:R(\nu)=R_{\min}\}$ is nonempty. Moreover, continuity of $R$ implies that $\Theta_0$
is closed in $\Theta$. Since a closed subset of a compact set
is compact, $\Theta_0$ is compact.

Next, choose any fixed $\nu_0\in\Theta_0$. The approximate-minimization
condition~\eqref{eq:optimization-gap} implies
\[
\widehat R_n(\widehat\nu_n)
\le
\inf_{\nu\in\Theta}\widehat R_n(\nu)+r_n
\le
\widehat R_n(\nu_0)+r_n.
\]
The definition of $\Delta_n$ bounds the difference between
empirical and population risks simultaneously for every
$\nu\in\Theta$. It therefore applies both to the fixed parameter
$\nu_0$ and to the data-dependent parameter $\widehat\nu_n$:
\[
R(\widehat\nu_n)-\widehat R_n(\widehat\nu_n)\le\Delta_n,
\qquad
\widehat R_n(\nu_0)-R(\nu_0)\le\Delta_n.
\]
Decomposing the population excess risk gives
\begin{align*}
R(\widehat\nu_n)-R_{\min}
&=
\left\{R(\widehat\nu_n)
      -\widehat R_n(\widehat\nu_n)\right\}+
\left\{\widehat R_n(\widehat\nu_n)
      -\widehat R_n(\nu_0)\right\}+
\left\{\widehat R_n(\nu_0)-R(\nu_0)\right\}\\
&\le \Delta_n+r_n+\Delta_n\\
&=2\Delta_n+r_n.
\end{align*}
Since $\widehat\nu_n\in\Theta$, its population risk cannot be
smaller than $R_{\min}$. Hence,
\[
0\le R(\widehat\nu_n)-R_{\min}\le2\Delta_n+r_n.
\]
In particular, for every $\eta>0$,
\begin{align*}
\mathbb P\left\{R(\widehat\nu_n)-R_{\min}>\eta\right\}
&\le
\mathbb P(2\Delta_n+r_n>\eta)\\
&\le
\mathbb P(\Delta_n>\eta/4)
+
\mathbb P(r_n>\eta/2)
\longrightarrow0.
\end{align*}
Thus, the population excess risk converges to zero in probability.

We now show that $\operatorname{dist}(\widehat\nu_n,\Theta_0)\xrightarrow{P}0.$ Fix $\varepsilon>0$ and consider
\[
\Theta_\varepsilon
:=
\left\{
\nu\in\Theta:
\operatorname{dist}(\nu,\Theta_0)\ge\varepsilon
\right\}.
\]
The distance function is continuous. Indeed, the triangle
inequality implies
\[
\operatorname{dist}(\nu,\Theta_0)
\le
\|\nu-\mu\|+\operatorname{dist}(\mu,\Theta_0),
\]
and exchanging $\nu$ and $\mu$ yields
$
\left|
\operatorname{dist}(\nu,\Theta_0)
-
\operatorname{dist}(\mu,\Theta_0)
\right|
\le\|\nu-\mu\|.
$
Therefore, $\Theta_\varepsilon$ is closed in $\Theta$ and hence
compact. If $\Theta_\varepsilon$ is empty, every parameter in $\Theta$
is within distance $\varepsilon$ of $\Theta_0$, so
\(
\mathbb P\left\{
\operatorname{dist}(\widehat\nu_n,\Theta_0)
\ge\varepsilon
\right\}=0.
\)
Suppose instead that $\Theta_\varepsilon$ is nonempty. Continuity
and compactness imply that the risk gap
\(
\kappa_\varepsilon
:=
\min_{\nu\in\Theta_\varepsilon}
\{R(\nu)-R_{\min}\}
\)
is attained and is nonnegative. In fact,
$\kappa_\varepsilon>0$. Otherwise, some
$\nu_\varepsilon\in\Theta_\varepsilon$ would satisfy
$R(\nu_\varepsilon)=R_{\min}$, implying
$\nu_\varepsilon\in\Theta_0$. This would give
\(
\operatorname{dist}(\nu_\varepsilon,\Theta_0)=0,
\)
contradicting its membership in $\Theta_\varepsilon$.

Consequently, whenever
$\operatorname{dist}(\widehat\nu_n,\Theta_0)\ge\varepsilon$,
we have
\[
R(\widehat\nu_n)-R_{\min}\ge\kappa_\varepsilon.
\]
Combining this implication with the excess-risk bound gives
\begin{align*}
\mathbb P\left\{
\operatorname{dist}(\widehat\nu_n,\Theta_0)
\ge\varepsilon
\right\}
&\le
\mathbb P\left\{
R(\widehat\nu_n)-R_{\min}\ge\kappa_\varepsilon
\right\}\\
&\le
\mathbb P(2\Delta_n+r_n\ge\kappa_\varepsilon)\\
&\le
\mathbb P(\Delta_n\ge\kappa_\varepsilon/4)
+
\mathbb P(r_n\ge\kappa_\varepsilon/2)
\longrightarrow0.
\end{align*}
The final convergence follows from
$\Delta_n\xrightarrow{P}0$, $r_n\xrightarrow{P}0$, and the
strictly positive, deterministic gap $\kappa_\varepsilon$.
Since $\varepsilon>0$ was arbitrary,
\[
\operatorname{dist}(\widehat\nu_n,\Theta_0)
\xrightarrow{P}0.
\]

Finally, if $\Theta_0=\{\nu_0\}$, then
\[
\operatorname{dist}(\widehat\nu_n,\Theta_0)
=
\|\widehat\nu_n-\nu_0\|,
\]
so $\widehat\nu_n\xrightarrow{P}\nu_0$.
When $\Theta_0$ contains multiple parameters, convergence to
the set alone does not imply convergence to any particular
member of that set.
\end{proof}

\subsection{Proof for Theorem~\ref{thm:dist}}
\begin{proof}
If $\nu_k\to\nu_0\in\Theta_0$, pathwise continuity gives $U_{\nu_k}(x,\epsilon)\to U_{\nu_0}(x,\epsilon)$ almost surely. At a continuity point of $Q_{0,x}$, dominated convergence of the indicators gives convergence of their survival probabilities. This convergence is uniform as the distance to $\Theta_0$ tends to zero: otherwise a sequence approaching $\Theta_0$ with survival error bounded away from zero has, by compactness, a subsequence converging to an element of $\Theta_0$, a contradiction. Apply the set convergence in~\eqref{eq:argmin-risk}.
\end{proof}

\subsection{Proof for Corollary~\ref{cor:T}}
\begin{proof}
Proposition~\ref{prop:target} gives $Q_{\nu_0,x}=P_x^*$ almost everywhere for each $\nu_0\in\Theta_0$. Intersect these full-measure sets over a countable dense subset of the compact set $\Theta_0$. By (A2) and Fubini, also restrict to a full-measure set of $x$ where parameter continuity holds for almost every noise value. Common-noise weak convergence extends equality of laws from the dense subset to every minimizer, giving the common set. Under residual agreement, Proposition~\ref{prop:pooling} gives $P_x^*=P_x$, and $1-F_x(\log t)=S_T(t\mid x)$ for $t<\tau$.

Conditional on the fitted model, the Dvoretzky--Kiefer--Wolfowitz inequality \citep{massart1990tight} gives, for every $\varepsilon>0$,
\[
\mathbb P\left(\sup_z\left|\widehat F_{n,M_n}(z\mid x)-F_{\widehat\nu_n,x}(z)\right|>\varepsilon
\,\middle|\,\widehat\nu_n\right)\le2e^{-2M_n\varepsilon^2}.
\]
Taking expectations and using the triangle inequality proves the Monte Carlo assertion.
\end{proof}

\subsection{Proof for Proposition~\ref{prop:excess}}
\begin{proof}
By~\eqref{eq:energy-identity}, subtracting the finite constant $\int r_x(P_x^*)P_X(dx)$ from $R(\nu)$ gives the integrated squared CDF error against $F_X^*$. Hence its infimum is $a_\Theta$, and~\eqref{eq:argmin-risk} proves~\eqref{eq:excess-pseudo}. Since CDF differences have magnitude at most one,
\[
\int(F_x^*-F_x)^2\,dz\le\int|F_x^*-F_x|\,dz=W_1(P_x^*,P_x)\le b(x).
\]
Use $(v+w)^2\le2v^2+2w^2$, integrate over $X$, and apply~\eqref{eq:excess-pseudo} to prove~\eqref{eq:true-cdf-bound}. On $(-\infty,a]$, exponentiation is $\tau$-Lipschitz, so a coupling from Proposition~\ref{prop:pooling} yields both bounds in~\eqref{eq:time-wasserstein}. 
\end{proof}

\section{Region-Level Covariates}\label{appendix:region_covariates}
To capture complex, unobserved institutional phenotypes that influence transplant outcomes, we engineered four region-level contextual features to train EnSurv. First, the leave-one-out delayed graft function (DGF) rate utilizes first-week dialysis requirements as a standard clinical proxy for DGF. We hypothesize that elevated DGF rates at a region reflect systemic vulnerabilities in organ procurement or perioperative care, which inherently shifts a region's baseline hazard for graft failure. To prevent data leakage, this rate was formulated for each patient by subtracting their individual DGF status from the total regional sum of DGF events, and dividing by the regional patient count minus one. Instances where a region contained only a single patient were imputed with the global DGF mean to ensure continuity. For the post-transplant setting, EnSurv assigns a negative weight to this variable, indicating that higher DGF is associated with accelerated graft failure.

Second, we quantified protocol rigidity through the Shannon entropy of region-specific immunosuppression regimens. We posit that low entropy indicates strict adherence to standard protocols, whereas high entropy reflects a region frequently customizing regimens to manage a highly complex, sensitized patient pool. This feature was formulated by concatenating six binary immunosuppression indicators into a single protocol signature string per patient. The Shannon entropy, calculated as $-\sum p \log_2(p)$ where $p$ represents the probability of each unique protocol signature within the region, was computed at the regional level and broadcast back to the individual patient records. EnSurv learns that a higher regional immuno-entropy is associated with longer graft survival.

Third, the region-level Patient Acuity Index was constructed to capture a region's aggregate risk appetite. We hypothesize that centers in regions frequently accepting marginal kidneys or highly sensitized patients possess fundamentally different baseline risks, which alters the expected survival probability of an otherwise average graft. Rather than a singular composite score, this index was operationalized by evaluating two parallel continuous variables: the region's mean KDRI to gauge organ marginality tolerance, and mean PRA score to capture highly sensitized patient volume. EnSurv assigns negative weights to both features, learning that higher regional average KDRI and PRA scores are negatively associated with graft outcomes in both pre- and post-transplant settings.

Finally, the Operational Strain Ratio serves as a measurement for infrastructural fatigue and complication burden, factors that may ultimately compromise post-discharge graft longevity. This was formulated by dividing the region's median length of stay by its volume capacity.

\section{Evaluation Metrics}\label{appendix:metrics}
Our evaluation framework encompasses a comprehensive suite of ten metrics, which we describe in this section.  Let $\mathcal D = \{(X_i, Y_i, \delta_i)\}_{i=1}^n$ represent the dataset, where $Y_i = \min(T_i, C_i)$ is the observed time, $T_i$ is the true event time, $C_i$ is the censoring time, and $\delta_i = \mathbf I(T_i \le C_i)$ is the event indicator. Let $\widehat{S}(t\vert{}X_i)$ be the predicted survival function, $\widehat{T}_i$ be the predicted expected or median survival time, and $\widehat{f}(X_i)$ be a general risk score.

\begin{itemize}
    \item \textbf{Harrell's Concordance index:} Harrell's C-index measures the proportion of correctly ordered comparable pairs, where a pair $(i,j)$ is comparable if $Y_i < Y_j$ and $\delta_i = 1$. The index is defined as \citep{harrell1996multivariable}: $$C = \frac{\sum_{i \neq j} \mathbf{1}(Y_i < Y_j, \delta_i = 1) \mathbf{1}\left(\widehat{f}(X_i) > \widehat{f}(X_j)\right)}{\sum_{i \neq j} \mathbf 1(Y_i < Y_j, \delta_i = 1)}.$$ It evaluates global ranking discrimination.

    \item \textbf{Antolini's Concordance index:}
    Antolini's C-index \citep{antolini2005time} is an extension of the concordance index for models predicting time-dependent survival probabilities. Concordance is evaluated at the specific observed event time of the index case rather than using a time-independent risk score: $$C_{td} = \frac{\sum_{i \neq j} \mathbf 1(Y_i < Y_j, \delta_i = 1) \mathbf 1\left(\widehat{S}(Y_i\vert{}X_i) < \widehat{S}(Y_i\vert{}X_j)\right)}{\sum_{i \neq j} \mathbf 1(Y_i < Y_j, \delta_i = 1)}.$$

    \item \textbf{Uno's Concordance index:} Harrell's C-index is strictly dependent on the study-specific censoring distribution, which leads to biased discrimination estimates when censoring is heavy. Uno's C-index \citep{uno2011c} provides a consistent, censoring-independent estimator by applying Inverse Probability of Censoring Weighting (IPCW). Let $\widehat{G}(t)$ denote the Kaplan-Meier estimator of the censoring distribution (the probability of not being censored up to time $t$). The metric evaluates concordance up to a pre-specified truncation time $\tau$, weighting each valid comparable pair by $\widehat{G}(Y_i)^{-2}$ to adjust for the probability that neither observation in the pair was censored prior to the index event time. 

    \item \textbf{Hinge error:} Hinge error is an absolute error metric that selectively penalizes point predictions for right-censored instances. For an observed event ($\delta_i=1$), the error is $\vert{}Y_i - \widehat{T}_i\vert{}$. For a censored observation ($\delta_i=0$), the error is penalized only if the predicted time is earlier than the censoring time, yielding $\max(0, Y_i - \widehat{T}_i)$. The mean Hinge Error is: $$\text{Hinge Error} = \frac{1}{n} \sum_{i=1}^n \big[ \delta_i \vert{}Y_i - \widehat{T}_i\vert{} + (1 - \delta_i) \max(0, Y_i - \widehat{T}_i) \big].$$

    \item \textbf{Margin error:} Margin error \citep{haider2020effective} is an adaptation of MAE that imputes the unobserved true event times for censored samples using KM expected values. Let $\mathbb E_{KM}[T_i \vert{} T_i > Y_i]$ be the conditional expected survival time given survival past the censoring time $Y_i$. The error is $\ell_i = \vert{}Y_i - \widehat{T}_i\vert{}$ if $\delta_i = 1$, and $\ell_i = \left\vert{}\mathbb E_{KM}[T_i \right\vert{} T_i > Y_i] - \widehat{T}_i\vert{}$ if $\delta_i = 0$. The overall error is the mean over all $n$ instances.

    \item \textbf{Uncensored MAE:} This is the standard mean absolute error computed exclusively on the uncensored subgroup $U = \{i \mid \delta_i = 1\}$, ignoring right-censored observations: $$\text{MAE}_{unc} = \frac{1}{\vert{}U\vert{}} \sum_{i \in U} \vert{}Y_i - \widehat{T}_i\vert{}.$$

    \item \textbf{AUC:} AUC denotes the Area Under the ROC Curve evaluated at a specific time horizon $t$, given by: 
    $$\text{AUC}(t) = \frac{\sum_{i=1}^n \sum_{j=1}^n \mathbf 1(Y_i \le t, \delta_i=1) \mathbf 1(Y_j > t) \mathbf 1\left(\widehat{f}(X_i) > \widehat{f}(X_j)\right)}{\sum_{i=1}^n \sum_{j=1}^n \mathbf 1(Y_i \le t, \delta_i=1) \mathbf 1(Y_j > t)}.$$

    \item \textbf{Brier score:} The time-dependent Brier score $BS(t)$ measures the mean squared error between the true survival status $\mathbf 1(Y_i > t)$ and the predicted probability $\widehat{S}(t\vert{}X_i)$, weighted by IPCW: $$BS(t) = \frac{1}{n} \sum_{i=1}^n \widehat{W}_i(t) \big(\widehat{S}(t\vert{}X_i) - \mathbf 1(Y_i > t)\big)^2.$$

    \item \textbf{IBS:} The Integrated Brier Score calculates the integral of the time-dependent Brier Score $BS(t)$ over a time range $[0, t_{\max}]$:
    $$\text{IBS} = \frac{1}{t_{\max}} \int_0^{t_{\max}} BS(t) dt.$$ Plain IBS ignores the IPCW-weighting. 

    \item \textbf{KM-Calibration:} This metric compares the average predicted survival curve of a model against the empirical Kaplan-Meier survival curve. Lower values indicate better alignment and calibration, where 0 represents a perfect match and higher values denote greater miscalibration.

\end{itemize}

All metrics are computed using the \texttt{SurvivalEVAL} library \citep{qi2023survivaleval}, which can be referred to for a more detailed description of the metrics.

\section{Quantitative Tables}\label{appendix:tables_figures}




\begingroup
\scriptsize
\setlength{\tabcolsep}{4pt}
\renewcommand{\arraystretch}{0.8}
\setlength{\LTcapwidth}{\textwidth}
\begin{longtable}{@{}llccc@{}}
\caption{Descriptive statistics and explanation for covariates in the UNOS dataset for kidney transplantation utilized in this study. Continuous variables are summarized as mean $\pm$ SD, and categorical variables are summarized as count (\%) within each event stratum.} \label{table:data_explanation}\\
\toprule
\textbf{Covariate} & \textbf{Description} & \makecell{\textbf{Graft Failure Observed}\\$(N=98140)$} & \makecell{\textbf{No Event/Censored}\\$(N=145303)$} & \textbf{\% Missing}
 \\

\midrule
\endfirsthead

\multicolumn{5}{c}{
\tablename\ \thetable\ -- continued from previous page
}\\
\toprule
\textbf{Covariate} & \textbf{Description} & \makecell{\textbf{Graft Failure Observed}\\$(N=98140)$} & \makecell{\textbf{No Event/Censored}\\$(N=145303)$} & \textbf{\% Missing} \\

\midrule
\endhead

\midrule
\multicolumn{5}{r}{Continued on next page}\\
\endfoot

\bottomrule
\endlastfoot

hcvdr & HCV Status (Donor/Recipient) & ~ & ~ & 0.0\% \\ 
        ~ & $\quad$ D+R+ & 2343 (2.4\%) & 2534 (1.7\%) & ~ \\ 
        ~ & $\quad$ D+R- & 662 (0.7\%) & 2834 (2.0\%) & ~ \\ 
        ~ & $\quad$ D-R+ & 4822 (4.9\%) & 4516 (3.1\%) & ~ \\ 
        ~ & $\quad$ D-R- & 90313 (92.0\%) & 135419 (93.2\%) & ~ \\ \midrule
        age\_n & Age at transplant (years) & 50.12 $\pm$ 15.44 & 49.36 $\pm$ 15.09 & 0.0\% \\ \midrule
        LOS & Recipient length of stay post transplant (days) & 9.19 $\pm$ 15.14 & 6.63 $\pm$ 8.89 & 16.0\% \\ \midrule
        bmi\_grouping & BMI Category & ~ & ~ & 0.0\% \\
\bottomrule

\end{longtable}
\endgroup

\begingroup
\tiny
\setlength{\tabcolsep}{2.4pt}
\renewcommand{\arraystretch}{1.2}
\setlength{\LTcapwidth}{\textwidth}
\setlength{\LTcapwidth}{\textwidth}
\begin{longtable}{@{}ccccccccccccccc@{}}
\caption{Results on simulated survival datasets. The \underline{\textbf{best}} and \textbf{second-best} results are highlighted.
The mean and standard deviation are obtained across 3-fold cross validation for each simulation setting.} \label{table:simulation_results}\\

\toprule
\rotatebox{90}{Distribution} & \rotatebox{90}{Risk Type} & \rotatebox{90}{Censored Param} & \rotatebox{90}{Actual Censoring} & \rotatebox{90}{\# Samples} & \rotatebox{90}{Model} &	\rotatebox{90}{Harrell's C}	& \rotatebox{90}{Antolini's C} &	\rotatebox{90}{Hinge Error} &	\rotatebox{90}{Margin Error} &	\rotatebox{90}{Uncensored MAE} & \rotatebox{90}{AUC} &	\rotatebox{90}{IBS} & \rotatebox{90}{AUPRC} &	\rotatebox{90}{KM-Calibration}
 \\

\midrule
\endfirsthead

\multicolumn{15}{c}{
\tablename\ \thetable\ -- continued from previous page
}\\
\toprule
\rotatebox{90}{Distribution} & \rotatebox{90}{Risk Type} & \rotatebox{90}{Censored Param} & \rotatebox{90}{Actual Censoring} & \rotatebox{90}{\# Samples} & \rotatebox{90}{Model} &	\rotatebox{90}{Harrell's C}	& \rotatebox{90}{Antolini's C} &	\rotatebox{90}{Hinge Error} &	\rotatebox{90}{Margin Error} &	\rotatebox{90}{Uncensored MAE} & \rotatebox{90}{AUC} &	\rotatebox{90}{IBS} & \rotatebox{90}{AUPRC} &	\rotatebox{90}{KM-Calibration} \\

\midrule
\endhead

\midrule
\multicolumn{15}{r}{Continued on next page}\\
\endfoot

\bottomrule
\endlastfoot

        Exponential & Linear & 5 & 59.22\% & 5000 & EnSurv & $\bms{0.9041}{0.0043}$ & $\ms{0.9039}{0.0045}$ & $\ubms{0.8343}{0.0303}$ & $\ubms{4.2651}{0.7785}$ & $\ms{1.3889}{0.0235}$ & $\bms{0.9555}{0.0046}$ & $\ubms{0.0321}{0.0031}$ & $\ms{0.6613}{0.0048}$ & $\ms{0.0101}{0.0023}$ \\ 
        ~ & ~ & ~ & ~ & ~ & CoxPH & $\ubms{0.9054}{0.0048}$ & $\bms{0.9053}{0.0048}$ & $\ms{1.4611}{0.6947}$ & $\ms{128.0743}{134.2426}$ & $\ms{3.2976}{1.7721}$ & $\ubms{0.9567}{0.0052}$ & $\bms{0.0328}{0.0039}$ & $\ms{0.6922}{0.0015}$ & $\ms{0.0058}{0.0073}$ \\ 
        ~ & ~ & ~ & ~ & ~ & WeibullAFT & $\ubms{0.9054}{0.0048}$ & $\ubms{0.9054}{0.0048}$ & $\ms{1.7487}{0.5550}$ & $\ms{363.1454}{170.4581}$ & $\ms{4.0046}{1.4268}$ & $\ubms{0.9567}{0.0052}$ & $\bms{0.0328}{0.0031}$ & $\bms{0.6999}{0.0024}$ & $\ms{0.0055}{0.0070}$ \\ 
        ~ & ~ & ~ & ~ & ~ & DeepHit & $\ms{0.5530}{0.0701}$ & $\ms{0.4891}{0.0041}$ & $\ms{2.9474}{0.0661}$ & $\ms{7.7346}{1.0315}$ & $\bms{1.3657}{0.0541}$ & $\ms{0.5729}{0.0290}$ & $\ms{0.1373}{0.0127}$ & $\ms{0.3498}{0.0029}$ & $\ms{0.2095}{0.0163}$ \\ 
        ~ & ~ & ~ & ~ & ~ & DeepSurv & $\ms{0.6263}{0.0560}$ & $\ms{0.6263}{0.0560}$ & $\ms{3.0493}{0.1323}$ & $\ms{7.5423}{0.9276}$ & $\ms{2.2676}{0.9391}$ & $\ms{0.6537}{0.0709}$ & $\ms{0.1050}{0.0065}$ & $\ms{0.4035}{0.0206}$ & $\ms{0.0750}{0.0546}$ \\ 
        ~ & ~ & ~ & ~ & ~ & RSF & $\ms{0.8370}{0.0048}$ & $\ms{0.8346}{0.0039}$ & $\ms{2.3504}{0.2780}$ & $\ms{6.2326}{0.2975}$ & $\ms{1.6446}{0.2623}$ & $\ms{0.8904}{0.0046}$ & $\ms{0.0789}{0.0049}$ & $\ms{0.4544}{0.0063}$ & $\ms{0.0344}{0.0036}$ \\
\bottomrule

\end{longtable}
\endgroup

\begin{table}[H]
\caption{Results for modeling the time to graft failure on 3 folds of the UNOS dataset under the pre-transplant setting. The \underline{\textbf{best}} and \textbf{second-best} results are highlighted.} \label{table:preTX_kidney_results}
\begin{adjustbox}{width=\textwidth}
\begin{tabular}{cccccccccccccccccc}
\toprule
Fold & Model & \makecell{Harrell's\\C-index} & \makecell{Antolini's\\C-index} & \makecell{Uno's\\C-index} & \makecell{Hinge\\Error} & \makecell{Margin\\Error} & \makecell{Uncensored\\MAE} & \makecell{Plain\\IBS} & \makecell{IBS-\\IPCW} & AUPRC & \makecell{KM-Calib-\\ration} & \makecell{AUC\\1-Yr} & \makecell{Brier\\1-Yr} & \makecell{Uno's C\\1-Yr} & \makecell{AUC\\5-Yr} & \makecell{Brier\\5-Yr} & \makecell{Uno's C\\5-Yr} \\ \midrule

1 & CoxPH & 0.6011 & 0.6011 & 0.5944 & 935.2385 & 2125.8188 & 1973.8846 & 0.0699 & \underline{$\mathbf{0.1604}$} & \underline{$\mathbf{0.6586}$} & \underline{$\mathbf{0.0001}$} & 0.6175 & $\mathbf{0.0739}$ & 0.6082 & 0.626 & \underline{$\mathbf{0.1917}$} & 0.6014 \\ 
        ~ & WeibullAFT & $\mathbf{0.6013}$ & $\mathbf{0.6013}$ & 0.5923 & 952.145 & 2145.8609 & 2009.196 & 0.0756 & 0.1657 & 0.6447 & 0.0033 & $\mathbf{0.6189}$ & $\mathbf{0.0739}$ & $\mathbf{0.6101}$ & 0.6235 & 0.1937 & 0.602 \\ 
        ~ & CoxNet & 0.6009 & 0.6009 & 0.5944 & 1929.5051 & 3202.5789 & 1939.5667 & 0.1821 & 0.3714 & 0.1288 & 0.2109 & 0.6175 & 0.8334 & 0.6079 & $\mathbf{0.6261}$ & 0.6508 & 0.6011 \\ 
        ~ & RSF & 0.5921 & 0.5946 & 0.5854 & 1927.6496 & 3200.8633 & $\mathbf{1937.9550}$ & 0.1704 & 0.3494 & 0.1509 & 0.1839 & 0.5989 & 0.7699 & 0.5979 & 0.5984 & 0.6048 & 0.5915 \\ 
        ~ & GBSA & 0.6012 & 0.6012 & 0.5934 & 1929.4628 & 3202.4604 & 1939.232 & 0.1823 & 0.3749 & 0.1281 & 0.208 & 0.608 & 0.8279 & 0.6026 & 0.6184 & 0.6508 & $\mathbf{0.6027}$ \\ 
        ~ & SurvivalTree & 0.5597 & 0.5603 & 0.5567 & 1934.281 & 3222.4728 & 1962.3718 & 0.1725 & 0.3527 & 0.1514 & 0.184 & 0.5621 & 0.7751 & 0.5626 & 0.5655 & 0.6084 & 0.5592 \\ 
        ~ & ExtraSurvivalTree & 0.5636 & 0.5632 & 0.563 & 1947.3599 & 3257.2331 & 1999.5714 & 0.1755 & 0.3571 & 0.1445 & 0.1919 & 0.5681 & 0.7934 & 0.5665 & 0.5732 & 0.6214 & 0.563 \\
\bottomrule

\end{tabular}
\end{adjustbox}
\end{table}

\end{appendices}
\end{document}